# Sirius-Tiamat: How Small Can a One-Pass Decision Encoder Get Before Its Decisions Change?

*Precision, layer deletion, weight rounding, early exit, distillation and vocabulary trimming of a 421-million-parameter typed-decision model, measured by flipped decisions on in-family, unseen-schema and out-of-family items, calibration drift and machine-independent cost*

Mohammed Nayeem Khan
mohammed.nayeem.khan@intel.com
mnknayeem@gmail.com

Vrishank Tiwari
vrishank.tiwari@intel.com

Intel Corporation

**Abstract**

A one-pass decision encoder reads a text, a question and several named options, and returns a probability for each option in a single forward pass. We ask how far Laya, a 421M-parameter encoder of this kind, can be compressed before its decisions change. We count decisions that differ from Laya's own full-precision (fp32) answer, and track calibration (how well confidence matches accuracy, measured by expected calibration error, ECE), on 1,221 synthetic items with the training schemas, 273 with an unseen schema, and 800 out-of-family questions from ARC, MMLU and HellaSwag. Layer pruning of a related model family is known; we add flips, calibration, held-out data and retraining. bf16 changes 0.2% of in-family decisions but 2.2% out of family; 6-bit grouped rounding 1.3% and 5.9%; deleting 6 layers 0.9% and 9.1%, while ECE rises from 5.9% to 39.2%. One fitted temperature, needing no labels, restores it to 3.6%. One scale per tensor fails below six bits, and CPU dynamic int8 changes 30% of decisions, nearly all through one layer. Total variation tracks flips for rounding but not for deletion, and an early exit saves layers over a static cut only if its confidence gate is first recalibrated. Distilling a 16-layer student cuts its in-family flips from 44% to 9% (45% out of family). At equal stored size, rounding flips fewer decisions than deletion. Data are synthetic and LLM-labelled, for one architecture.

**Key findings (every number recomputed at build time)**

1. In-family numbers understate flips. On 800 public-benchmark questions the same recipes change several times more decisions than on the 1,221 in-family items: bf16 0.2% against 2.2%, 4-bit grouped rounding 4.3% against 21.4% (Section 12).
2. Half precision is nearly free on the nine synthetic sets: the worst of bf16/fp16 changes 14 of 1,494 decisions on either checkpoint. The scale granularity of weight rounding matters more than the bit width: per-tensor 4-bit changes 76% of decisions, per-channel 9.6% and groups of 32 5.5%.
3. PyTorch dynamic int8 changes 30% of decisions; keeping the feed-forward output projection in fp32 brings it to 2.9% (Section 5). Community int8 exports should be checked the same way.
4. Layers can be deleted from the top without retraining (top 6 within 1% flips, top 9 within 5% in family), but the probabilities move first: ECE 5.9% to 39.2% at six layers. One temperature fitted without labels brings it to 3.6%, with labels 2.8% (Section 16); on the unseen schema it stays at 7.1% without labels and 14.8% with them.
5. Total variation tracks flips for weight rounding (median flips per unit of TV 0.92) and not for layer deletion; margins shift by an offset and not by shrinkage; the fp32 margin ranks flipped items with AUROC 0.91 for 4-bit grouped rounding (Section 14).
6. A confidence-gated early exit saves layers over static deletion only with a recalibrated gate: 20.4 layers at 1% in-family flips against 22 for the static cut, and 22.8 with the raw top probability (Section 15).
7. A second backbone (mmBERT-base, 22 layers) behaves like the main one under rounding (4-bit grouped 11.0% against 13.2% of decisions) but not under layer deletion (six layers: 3.7% against 21.4%), so the deletable depth has to be measured per model (Section 13).
8. A two-epoch distillation of the 12-layer student cuts in-family flips from 44% to 9% and out-of-family flips from 62% to 45% (Section 18).
9. At 600 MiB the best rounding configuration changes 1.5% of decisions and the best layer deletion 11.9%; the smallest tier-A recipe is 8-bit channel (bf16 weights) at 453 MiB (Section 19).
10. A per-task TF-IDF classifier scores 89.9% against 84.2% for fine-tuned Laya on six shared tasks (two out of distribution for Laya), at about 2.3 MiB and 1.2 ms per item.
11. Not done: GPU timings, idle-machine CPU timings, quantisation-aware training and layer dropout, and any real (non-synthetic) deployment data. All labels are LLM-made and all synthetic items come from one generator.

## 1 Summary

A decision encoder that answers in one pass is attractive because it is cheap. This study asks how much cheaper it can be made by changing the model alone (fewer bits, fewer layers, narrower layers, fewer vocabulary rows, a short retraining) before the decisions it returns differ from the uncompressed model's. The model is Laya, a 421-million-parameter encoder that reads a text, a question and named options and returns a probability for each option. We count *decision flips* against the model's own fp32 answer rather than only accuracy against labels, on 1,494 synthetic items in nine sets (eight from the training family of the fine-tuned checkpoint, one with an unseen schema) and on 800 public-benchmark questions that are out of family (Section 12).

**What is and is not new.** Training-free, decision-guided layer pruning of a decision model has been reported for Jev models [17], flips as a metric for compressed language models come from Dutta et al. [19], and the Hugging Face hub already lists at least nine community int8 builds of Laya without an evaluation (Section 2). We do not claim to be first to compress a decision model. FastJEV measures task-score retention of a multimodal model without retraining. We measure which decisions change on a text encoder, on items from inside and outside its training family, and we add early exit, temperature repair, distillation and a noise floor (Section 2 lists what each cited paper did and did not measure).

The headline is a shape, and the in-family numbers understate it. Some knobs are close to free in family (half precision, per-channel 8-bit weights), some cost a predictable amount per step (bits below six, layers beyond 6), and some are traps (per-tensor scales below six bits, dynamic int8 on the feed-forward output projection, deleting low layers, slicing feed-forward units or heads). On the public-benchmark questions every recipe changes at least as large a share of decisions: bf16 changes 0.2% in family and 2.2% out of family, 6-bit grouped rounding 1.3% and 5.9%, deleting six layers 0.9% and 9.1% (Section 12). Deleting top layers is the surprising case because decisions survive while the reported probabilities do not, and a single temperature repairs the probabilities without labels (Section 16). A two-epoch distillation of a 16-layer student cuts its in-family flips from 44% to 9% but leaves 45% of out-of-family decisions changed (Section 18). At the same stored size, weight rounding flips fewer decisions than deleting layers (Section 19). Against a classical baseline, TF-IDF with logistic regression is within noise of the compressed model on several tasks at a tiny fraction of the cost (Section 20).

**Table 1.** Predictions stated before each experiment (the log is in the project's review folder) and what the data show. A prediction marked failed is reported in the text of the section named.

| Prediction | Result | Verdict |
|---|---|---|
| Flips on public-benchmark rows are at least as high as in family, at every setting | 35 of 35 configurations | held |
| Flips per unit of mean total variation: about 1 in family, below 0.6 out of family | median 0.81 in family (0.92 for rounding), 1.75 out of family | failed (out of family) |
| A linear margin fit ($R^2$ above 0.7) predicts the flip rate of held-out items within 2 points | rounding: $R^2$ 0.72 to 1.00, error up to 1.1 points; deletion, slicing and sparsity: $R^2$ 0.26 to 0.83, error up to 9.0 | held for rounding, failed otherwise |
| A confidence-gated early exit uses fewer layers than static deletion at 1% flips | raw gate 22.8 layers, recalibrated gate 20.4, static deletion 22 | held with a recalibrated gate, failed with the raw gate |
| One fitted temperature brings the six-layer ECE under 15% | 39.2% to 2.8% (synthetic sets); unseen schema 14.8% | held |
| Distillation brings in-family flips at 12 deleted layers under 10% and beats labels | distilled 8.6%, labels 9.7%, untrained 44.0% | held |
| The ordering of recipes by flip rate is the same on other checkpoints (Kendall tau above 0.8) | tau 0.78 to 1.00 over 4 checkpoints | failed for other backbone (mmBERT-base) |
| Rerunning fp32 with other batch sizes or row order flips at most 0.2% of decisions | at most 0 of 2,294 (0.00%) | held |
| At matched stored size, rounding flips fewer decisions than layer deletion, slicing or sparsity | at 600 MiB: rounding 1.5%, deletion 11.9% | held |
| The fp32 margin alone predicts which items flip under 4-bit grouped rounding (AUROC above 0.85) | AUROC 0.91 | held |

All labels are written by language models, the synthetic data come from one generator, the public-benchmark questions are a stress test for an encoder trained on other schemas and not a benchmark of its skill, all conclusions are about a handful of checkpoints of one architecture, and the machine was shared: machine-independent cost (parameters, bytes, FLOPs) is the main cost axis and CPU timings are noisy context. Section 21 says what that does and does not license.

## 2 Related work and what is new

Each of the 30 works cited here was looked up on 10 and 18 November 2026 at its Hugging Face paper page or arXiv abstract page to confirm that title, authors and identifier match. For four of them (FastJEV, Dutta et al., Almeida Felicio, and Blagojevich and Sunderland) we also read the introduction and the description of the experiments in the downloaded text. For the others we read the abstract and, for the encoder paper, the architecture description, so what we say about those is limited to what their abstracts claim. The reference list is at the end of the report.

### 2.1 Typed decision models and their cost

Typed decision models answer a bounded question from one forward pass and are pitched as a cheap replacement for calling a language model. Li et al. [6] report 22.7 to 64.5% lower median decision latency than the fastest of three hosted LLMs for a Jev decision model on an edge admission path, and Cheng et al. [7] describe a 2-billion-parameter typed decision model that answers in about 31 ms on a laptop GPU. TypedBench [8] argues that such models must be judged jointly on policy adherence, wording robustness, probability quality and induced decisions. We adopt its emphasis on probability quality (Section 11) but use plain equal-width ECE without a noise floor, so our calibration numbers are not comparable with theirs. Rafe and Das [16] compare eight decision-model checkpoints with trained classifiers and generative models on accuracy; size reduction is outside their abstract. These papers treat the model as a fixed artefact. We vary the artefact itself.

### 2.2 The closest prior work: compression of decision models

Two papers sit nearest to this one, and a third kind of evidence is not a paper at all. **FastJEV** [17] targets Jev decision models that score candidate decisions for multimodal inputs. Besides reusing recurrent states and sharing candidate prefixes, it prunes depth with a training-free rule that looks at relative score changes on a small unlabelled set. By its abstract, the complete method cuts candidate depth by 43.75 to 45.83% across three OmniJev sizes and keeps 93.66 to 97.52% of the original task score on average over six evaluation sets. In the paper's experiments the layer ranking comes from 32 OK-VQA inputs, the same mask is reused on every task, the shared prefix still runs every block, and the outcome is accuracy retention against the original model on public multimodal benchmarks and a reconstructed robotics question set. That is the same idea as our greedy layer deletion (Section 7), applied to a different model family, a different input modality and a different outcome (task score, not the share of changed decisions). It has no retraining, no calibration measurement and no evaluation on a schema or a domain outside the benchmarks. We do not claim layer deletion for a decision model as new. We keep the part that is ours and that FastJEV does not touch: which decisions change (not only how much accuracy is kept), how the reported probabilities drift, what happens on items from outside the training family, how layer deletion compares with rounding, slicing, early exit and distillation at the same size, and whether a short retraining recovers the loss (Section 18). We report our retained-score numbers next to theirs wherever the quantities are comparable (Section 7), and we say where they are not.

**Stekel** [18] asks how small an encoder for closed-set decisions can be, comparing joint and per-option reading at 17 million to 1 billion parameters, and reports that the smallest released joint encoder answers a median request in 14 ms on 16 processor threads. Its route to a small model is to pick a smaller trained model, with accuracy and latency as the outcomes. Ours is to shrink one trained model after the fact and to ask which decisions change. The two are complementary: that paper gives the accuracy-versus-size curve across independently trained models, this one gives the curve inside a single model and adds flips and calibration drift. It also reports that reordering options changes 12 to 23% of the answers of joint models, a reminder that one-pass decision encoders have a flip problem even before any compression.

**Community builds.** On 10 November 2026 a search of the Hugging Face hub for Laya-related repositories returned nine whose names contain 'int8' (ONNX, OpenVINO and MNN exports, English and multilingual), more than thirty ONNX conversions, and single fp16 and int4 variants. We read repository names and creation dates only, not the model cards, and found no accompanying paper or evaluation of changed decisions. Quantised Laya is therefore already in use; what we add is a measurement of what quantising it, and the other compression knobs, does to its decisions.

### 2.3 Compression of transformer encoders

The tools we use are standard. Dettmers et al. [2] showed that int8 matrix multiplication of large language models needs special handling of a few outlier feature dimensions once models pass several billion parameters; our encoder is a thousand times smaller than that regime, so we test rather than assume where plain rounding breaks. Sajjad et al. [3] dropped layers from BERT, RoBERTa and XLNet and kept up to 98% of GLUE performance at 40% fewer layers, with the lower layers the most critical, a pattern we also see; Fan et al. [4] train with structured layer dropout so that shallower sub-networks can be selected at inference time without fine-tuning, and DistilBERT [5] obtains a smaller, faster student by distillation. Q8BERT [9] trains BERT with quantisation-aware fine-tuning for a 4x size reduction at little accuracy cost; we only round after training. D K et al. [15] prune the last layers of large language models used as text encoders before a short supervised run and report that up to 30% of layers go with negligible loss, which parallels our top-layer deletion (they retrain briefly, we do not). Vocabulary trimming [10] deletes embedding rows to shrink multilingual models; we do the same on an English-heavy decision encoder and add a test of what happens to the decisions when a test token falls outside the kept list.

### 2.4 Flips, divergence and noise

Dutta et al. [19] introduced 'flips' for compressed language models, the share of answers that change although accuracy does not, and measured them on benchmarks including MMLU, HellaSwag and ARC; we use their metric and, in Section 12, the same benchmarks, converted to one-pass decisions. Almeida Felicio [20] reports that the flip rate is close to the mean total variation between the dense and compressed distributions for 19 language models, which we test on an encoder in Section 14; Wu et al. [21] propose that quantization shrinks decision margins roughly linearly, which we test in the same section. Blagojevich and Sunderland [22] state each quantization effect as the excess over the flips that engine, restart and batch changes produce on two decision models; Section 13 measures the corresponding floor for Laya. None of these measures layer deletion, slicing or retraining on a typed decision encoder, and none of them reports out-of-family items for the decision model itself. Rababah et al. (2607.08734, abstract only) report a related 'correctness agreement' metric for quantised language models; we did not use it. Azizi et al. [29] (abstract only) report that deleting every option definition leaves the accuracy of an open typed decision model, laya-td among them, almost unchanged, so the option labels carry much of the decision; on our public-benchmark rows the labels are bare letters and only the descriptions carry the answer, which may be part of why the model is near chance there.

### 2.5 Healing, early exit and the choice between pruning and quantization

Healing a pruned model with a short training run is standard for language models: Gromov et al. [24] remove deep layers and heal with low-rank adapters, and ShortGPT [25] scores layers by how little they change the hidden state; Kim et al. [30] report that the calibration data matter more than the search algorithm for which layers are removed. Patient distillation [26] trains a shallow student on a deep teacher's intermediate layers. Early exit and cascades spend fewer layers on easy inputs [27] [28]. Kuzmin et al. [23] compare pruning with quantization on accuracy and find quantization better in most settings. We run each of these on one decision encoder and score them with the same outcome, decisions changed, on in-family, unseen-schema and out-of-family items (Section 15, Section 18, Section 19).

### 2.6 Who loses when a model is compressed

Aggregate accuracy hides changes in individual predictions. Hooker et al. [11] show that compressed networks diverge on a narrow subset of examples even when top-line accuracy matches, which is the motivation for counting decision flips against the full-precision model, as we do. Proskurina et al. [12] find that 4-bit GPTQ quantisation lowers confidence in true labels for language models and that it disproportionately affects samples on which the full model was already unsure, which is the pattern Figure 11 shows for flips here, and Guo et al. [14] supply the calibration vocabulary (ECE, reliability) that we use for the drift measurements. Jiang et al. [13] call the flip rate between a deployed and a new model 'churn' and treat it as a cost in itself; a compressed model replacing a full one in a software branch has the same exposure.

The Laya checkpoints were not trained with layer dropout or distillation, so our layer-deletion results are the 'naive' setting on a different architecture (ModernBERT [1] mixes windowed and global attention).

### 2.7 What is, and is not, new here

- **Not new:** that decision-guided or training-free layer pruning of a decision model works (FastJEV), that compact encoders can serve closed-set decisions (Stekel), that quantised Laya runs (the community builds), that flips are a better view of compression damage than accuracy (Dutta et al.), that flips follow total variation for rounding (Almeida Felicio), that decision margins explain which items flip (Wu et al.), that a quantization comparison needs a noise floor (Blagojevich and Sunderland), that bf16/fp16 are nearly free, that per-channel and grouped scales beat a per-tensor scale at low bit width, that quantization beats pruning at equal size (Kuzmin et al.), that upper layers of an encoder are more expendable than lower ones, that unstructured sparsity does not speed up dense kernels, and that a small vocabulary frees embedding memory. Those are known results; we reproduce or test them on this model.
- **New, as far as we can tell:** (i) all of the above measured on one typed decision encoder with decision flips against the model's own fp32 answer as the outcome and accuracy secondary; (ii) the same grid on in-family items, on an unseen schema and on 800 out-of-family public-benchmark questions, which shows that the in-family numbers understate flips several-fold (Section 12); (iii) a test of the total-variation and margin accounts on an encoder, where total variation holds for rounding and fails for layer deletion, and margins shift by an offset and not by shrinkage (Section 14); (iv) a calibration repair for layer-deleted models with one temperature that needs no labels (Section 16); (v) early exit, distillation and label training compared with static deletion at the same depth (Section 15, Section 18); (vi) a noise floor and a replication on other checkpoints (Section 13); (vii) a recipe card with intervals and matched-size comparisons (Section 19); (viii) a comparison with a TF-IDF classifier on cost as well as accuracy.
- **Not claimed:** any new compression algorithm; results for models other than these checkpoints; results on real (non-synthetic) deployment data; anything about quantisation-aware training, which we did not try.

*Search*. On 10 November 2026 we queried the Hugging Face paper search with fourteen phrases, among them 'typed decision model quantization', 'Jev decision model compression', 'typed decision model layer pruning', 'quantization BERT classifier decision flips', 'vocabulary trimming embedding pruning', 'prediction churn model compression flips' and 'compression disproportionately impacts', and read the abstracts of the relevant hits. On 18 November we ran twenty more, on flips and quantization, layer-pruning calibration, early exit and cascades, healing after layer removal, noise in quantization comparisons and held-out typed-decision schemas, and read forty further abstracts (logs in the project's review folder). We found typed-decision papers on benchmarking, wording sensitivity, label bias, calibration and latency, general compression papers, FastJEV and the compact-encoder study above, and the flip, noise-floor and margin papers cited in Section 2. We found none that measures changed decisions, calibration drift, early exit and retraining together on one typed decision encoder, and none that evaluates a compressed typed decision model out of family. A title-level search of one index is weak evidence; a paper on this topic may exist.

## 3 The model being shrunk

Laya is a *typed decision model*: it is shown a piece of text (the state) and one or more typed questions, pick one of several named options, rate on an ordinal scale, or answer yes/no, and it returns a probability for every option. Nothing is generated: the question, its options and the state are laid out in one input sequence, with a mask slot in front of every option, and a single forward pass of an encoder reads the hidden state at each slot. The released checkpoint builds on the ModernBERT-large encoder [1]: 28 transformer layers of width 1,024 with GeGLU feed-forward blocks, 16 attention heads, rotary positions, and attention that is windowed (128 tokens) in two of every three layers and global in layers 0, 3, 6, 9, 12, 15, 18, 21, 24, 27. On top of the encoder sit a two-layer transformer 'option head' that mixes the option slots, a scorer that turns each slot into one logit, and a small act head. A temperature per question type converts logits to probabilities. Table 3 lists the parameters; the encoder holds 81% of them and the shipped file stores them in half precision (804 MiB).

**Table 3.** Where the 421,293,830 parameters of the fine-tuned checkpoint sit (counted from the safetensors header; the base checkpoint has identical tensor shapes).

| Tensor group | Parameters | Share |
|---|---:|---:|
| token embeddings (50,368 × 1,024) | 51,576,832 | 12.2% |
| attention QKV projections, 28 layers | 88,080,384 | 20.9% |
| attention output projections | 29,360,128 | 7.0% |
| FFN input (GeGLU, gate and value), 28 layers | 150,470,656 | 35.7% |
| FFN output projections | 75,235,328 | 17.9% |
| option head (two transformer layers) | 25,192,448 | 6.0% |
| scorer | 1,052,673 | 0.2% |
| act head | 263,938 | 0.1% |
| layer norms, temperature, other | 61,443 | 0.0% |

### 3.1 Two checkpoints, one training recipe

We shrink two checkpoints. **Base** is the published checkpoint with no task training. **Fine-tuned (v1)** is the base checkpoint trained for four epochs on 2,818 synthetic cases with the trainer that ships with the model (its default 'rlcd' objective, option-order shuffling for choice questions, learning rate 2.5e-5 for the encoder and 1e-4 for the heads, effective batch 64, then a short calibration and abstention-threshold fit on a held-back slice); one such run takes about five minutes on the 24 GB GPU used here. The training cases cover support-ticket routing, sentiment, ticket state serialised as JSON, hardware-validation failure classes, many-option topic choice, urgency on a three-level scale, and several non-English languages. Comparing the two checkpoints matters for this study because a model that is more confident and more accurate may be more or less fragile under shrinkage than one that is neither.

### 3.2 Where the data come from

All cases are synthetic. A 27-billion-parameter generator model (Qwen3.8-27B) was asked to write cases conditioned on a target label and category; the same model was then shown each case *without* the intended label and asked to label it blind, and every case on which the two passes disagreed was discarded. Test cases were generated separately from training cases and de-duplicated against them. An independent labeller (gemma4-31b) re-labelled the held-out items blind: it agrees with the stored label on 94%-99% of items depending on the set (Table 2). That is a consistency check, not ground truth: a label an LLM can reproduce is not necessarily one a human would give. For this study the labels are only a yardstick; the main quantity is whether a compressed model gives the *same decision as the uncompressed one*, which needs no labels at all.

**Table 2.** The nine held-out sets (1,494 items in total). 'Options' is the number of answer options per question; 'labeller agreement' is how often an independent LLM (gemma4-31b) gave the stored label when asked blind.

| Set | Items | Question types | Options | Labeller agreement |
|---|---|---|---|---|
| mixed probe | 224 | choice/noul/score | 2-12 | 97.3% |
| sentiment | 106 | choice | 3 | 99.1% |
| JSON tickets | 58 | choice | 3 | 96.6% |
| hardware failure | 82 | choice | 5 | 95.1% |
| 12-way topic | 195 | choice | 12 | 94.4% |
| multilingual | 322 | choice | 3 | 97.2% |
| urgency (ordinal) | 78 | score | 3 | 94.9% |
| code-mixed | 156 | choice | 3 | n/a |
| HR (unseen schema) | 273 | choice | 12 | 96.7% |

The nine sets are used exactly as they were held out for the fine-tuned model; all sixteen-item batches in every experiment are built from identical rows in identical order, so any two configurations can be compared item by item.

## 4 Knobs, measures and protocol

Six compression knobs are varied. Each is applied *in memory* to a pristine copy of the weights; nothing is written over an existing checkpoint. Where a physically smaller checkpoint is needed (Section 18) it is written under its own directory.

### 4.1 The six knobs

- **Precision (C).** fp32 with autocast disabled is the reference. We compare it with bf16 autocast (the runtime default), fp16 autocast, and fp16 or bf16 weights without autocast. On a CPU we also run PyTorch dynamic int8 (int8 weights, activations quantised on the fly), per tensor and per output channel, with and without the feed-forward output projections and attention.
- **Weight rounding (D).** Symmetric round-to-nearest to 8, 6, 5, 4 or 3 bits, dequantised back to floating point, so the effect on *decisions* is exact although no integer kernel is used. One scale per tensor, per output channel, or per group of 32 inputs; scope is the encoder linear layers, the encoder plus option head, or everything. Stored size counts packed integers plus fp16 scales.
- **Layer deletion (A).** Remove the top *k*, the bottom *k*, a middle block, every other, third or fourth layer, or the set chosen by a greedy search that deletes, at each step, the layer whose removal moves the answer probabilities least on 240 held-in training rows (never touching the test sets).
- **Slicing and sparsity (B).** Keep the top fraction of feed-forward units or attention heads ranked by a weight-norm product, or zero the smallest weights of every encoder matrix (unstructured sparsity saves no time on dense kernels and is reported for completeness).
- **Vocabulary (E).** The embedding table has 50,368 rows of 1,024 values. We replace rows never seen in the training texts by the unknown-token row and count the memory this would release.
- **Retraining.** Everything above is training-free. Section 18 adds a two-epoch re-fine-tune of layer-deleted students, on labels or on the fp32 model's own output distribution, and Section 15 tests early exit at the same depths.

### 4.2 What we count

**Decision flip.** An item flips when the top option of the compressed model differs from the top option of the uncompressed fp32 model *of the same checkpoint*. Flip rate is reported with a Wilson 95% interval over the 1,494 items; because flips are rare in the benign regime we also give raw counts. Accuracy against the stored labels is reported next to it but is the weaker signal: a model that changes 3% of its answers can keep its accuracy if the changes go in both directions. **Calibration drift** is the change in expected calibration error (10 equal-width bins on the top-option probability) relative to fp32, with a paired bootstrap interval (items resampled together). **Answer shift** (TV) is half the L1 distance between the two probability vectors, averaged over items; it flags small changes that do not flip anything. **Margin** is the log-probability of the fp32 model's top option minus that of the best other option; it is used in Section 14. **Item groups.** The nine synthetic sets split into eight in-family sets and one unseen-schema set, and Section 12 adds 800 public-benchmark questions; flips, accuracy and ECE are reported per group wherever the sections after Section 11 need it. **Noise floor.** Section 13 reruns fp32 with other batch sizes and row orders to show how many flips appear with nothing compressed.

**Cost.** Parameters, stored bytes and FLOPs per item are computed from tensor shapes and the mean test length; they are the main cost axis. CPU time was measured per configuration in a fresh process on 246 sentiment, hardware and JSON items, at batch 1 and 16 with a warm-up pass, and we report the median per-item time, the inter-quartile range and the CPU load sampled before each run. Other jobs ran on the machine during all of it, so these timings are noisy context, one run per configuration. No GPU timings were taken. Peak memory is the peak working set of the process.

**Machine.** Intel Core i9-13900K (24 cores, 32 threads), 64 GB RAM, one RTX 4090 with 24 GB shared with other jobs, Windows 10, PyTorch 2.11.0+cu128, `laya` 0.4.26. CPU timings use 8 threads. The accuracy sweeps ran on the GPU with fp32 matrix multiplies (autocast off) for every configuration that is not itself a precision test.

## 5 Precision: half formats and int8 on a CPU

The runtime's default is bf16 autocast, so the first question is what the reference itself hides. We compare four half-precision settings against fp32 on the same items (Table 4, Figure 1). For the fine-tuned checkpoint the worst of the four changes 3 of 1,494 decisions and for the base checkpoint 14; the single largest count is 14 (bf16 weights, no autocast, base checkpoint). None of these differences in accuracy between a half-precision setting and fp32 exceeds 0.13 points, and in every case the paired McNemar test against fp32 accuracy has $p \geq 0.50$. In practice, switching a deployment from fp32 to either 16-bit format is not a decision-changing act on these data; we make no stronger claim than that, because 14 flips out of 1,494 could still matter if the flipped items were the important ones (Section 10 looks at which items flip).

**Table 4.** Half-precision formats against fp32 on the same 1,494 items. Accuracy is against the stored labels; 'changed' counts items whose top option differs from fp32; ΔECE is in percentage points with a paired-bootstrap interval; TV is the mean answer shift in percent.

| Configuration | Accuracy [95% CI] | Decisions changed (count, rate, CI) | ΔECE (points) | TV (%) |
|---|---|---|---|---|
| fine-tuned: bf16 autocast (runtime default) | 87.3 [85.6-88.9] | 3 (0.20%) [0.07-0.59] | +0.13 [-0.57, +0.63] | 0.27 |
| fine-tuned: fp16 autocast | 87.3 [85.6-88.9] | 1 (0.07%) [0.01-0.38] | -0.06 [-0.20, +0.19] | 0.04 |
| fine-tuned: fp16 weights, no autocast | 87.3 [85.6-88.9] | 2 (0.13%) [0.04-0.49] | -0.12 [-0.50, +0.19] | 0.05 |
| fine-tuned: bf16 weights, no autocast | 87.4 [85.6-89.0] | 3 (0.20%) [0.07-0.59] | -0.00 [-0.60, +0.55] | 0.38 |
| base: bf16 autocast (runtime default) | 73.4 [71.1-75.6] | 6 (0.40%) [0.18-0.87] | +0.40 [-0.75, +1.35] | 0.37 |
| base: fp16 autocast | 73.5 [71.2-75.7] | 1 (0.07%) [0.01-0.38] | -0.00 [-0.38, +0.23] | 0.05 |
| base: fp16 weights, no autocast | 73.4 [71.1-75.6] | 2 (0.13%) [0.04-0.49] | -0.06 [-0.62, +0.34] | 0.07 |

| Configuration | Accuracy [95% CI] | Decisions changed (count, rate, CI) | ΔECE (points) | TV (%) |
|---|---|---|---|---|
| base: bf16 weights, no autocast | 73.6 [71.3-75.7] | 14 (0.94%) [0.56-1.57] | -0.48 [-1.33, +0.70] | 0.50 |

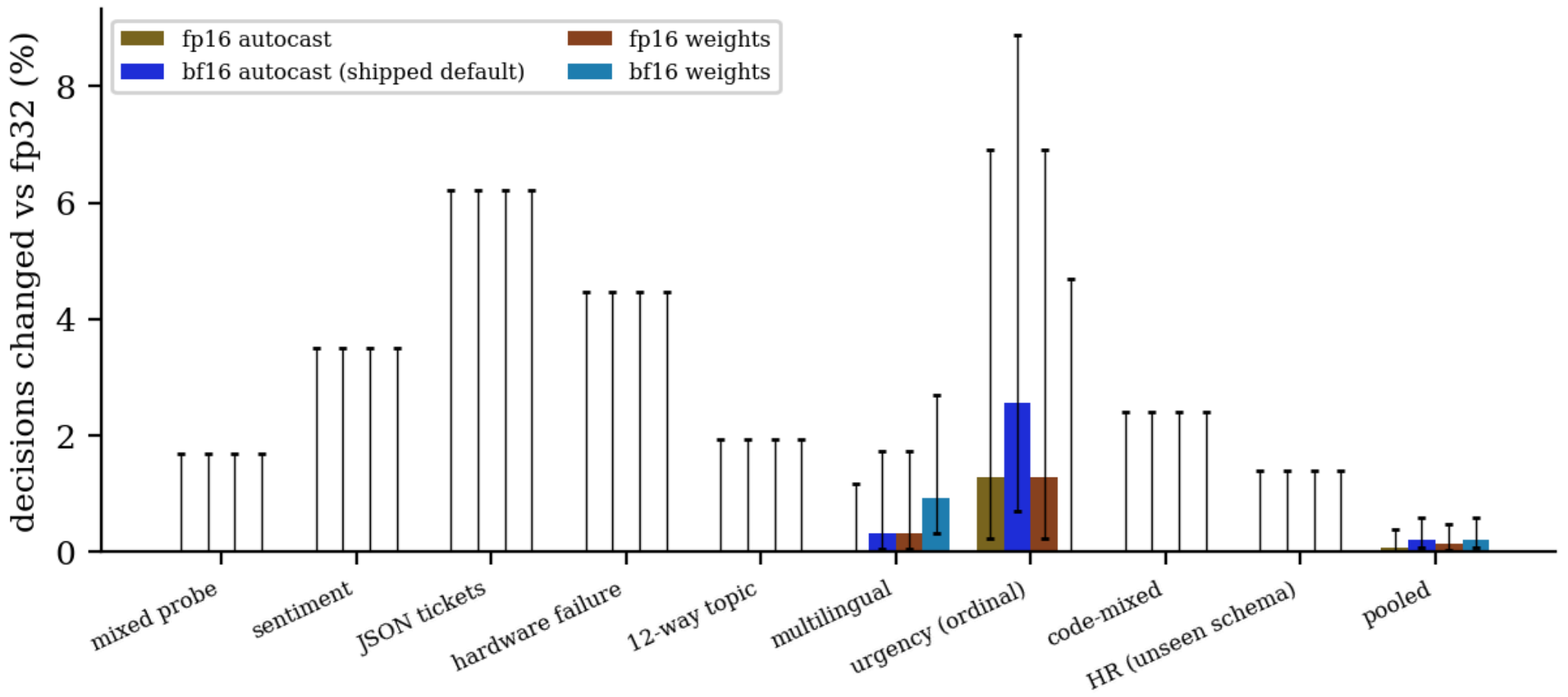

**Figure 1.** Share of decisions changed by each half-precision setting, per test set and pooled (fine-tuned checkpoint), with Wilson 95% intervals. Sets with fewer items have wide intervals.

**Dynamic int8 on a CPU.** PyTorch dynamic int8 stores weights as int8 and also quantises each linear layer's *input activations* on the fly, which weight-only rounding (Section 6) does not. Table 5 shows that this is where the damage is. The CPU's own fp32 run differs from the GPU fp32 run on 0 of 1,494 items, so the reference floor is 0. Quantising every encoder linear layer changes 454 items (30.4%) with a per-tensor weight scale and 450 (30.1%) with a per-channel one, and accuracy falls from 87.3% to 67.5%. Keeping only the feed-forward output projection in fp32 brings that to 44 items (2.9%); keeping attention in fp32 instead changes almost nothing (432 items), and quantising only the feed-forward input projection changes 28 (1.9%). The input to that output projection is the product of the GeGLU gate and value, a tensor whose few large-magnitude channels are the kind of outlier that Dettmers et al. [2] found to break plain 8-bit activation quantisation. We did not inspect those activations, so the attribution is a hypothesis that the ablation supports, not a demonstration. It means that a dynamic-int8 export of Laya, which is what several community builds are likely to be, should be checked for flipped decisions before it replaces the fp32 model; we tested PyTorch's implementation, not those builds.

**Table 5.** Dynamic int8 on the CPU (fine-tuned checkpoint). 'MiB' is the weight storage if written in int8 plus fp16 for everything left in floating point; 'ms' is the median time per item at batch 1 / batch 16 with 8 threads (see Section 17 for the noise).

| Configuration | Accuracy [95% CI] | Decisions changed vs GPU fp32 | ΔECE (points) | MiB | ms/item (b1 / b16) |
|---|---|---|---|---|---|
| CPU fp32 (no quantisation) vs GPU fp32 | 87.3 [85.5-88.9] | 0 (0.00%) [0.00-0.26] | +0.00 [-0.00, +0.00] | n/a | 132 / 114 |
| int8 dynamic, per-tensor scale | 67.1 [64.7-69.5] | 454 (30.39%) [28.11-32.77] | +5.34 [+3.48, +7.82] | 452 | 80 / 57 |
| int8 dynamic, per-channel scale | 67.5 [65.1-69.9] | 450 (30.12%) [27.85-32.50] | +4.98 [+2.52, +7.39] | 452 | 91 / 58 |
| per-tensor, FFN output kept fp32 | 86.6 [84.8-88.2] | 61 (4.08%) [3.19-5.21] | +0.15 [-1.08, +1.09] | 524 | 88 / 74 |
| per-channel, FFN output kept fp32 | 86.9 [85.1-88.6] | 44 (2.95%) [2.20-3.93] | -0.22 [-1.28, +0.69] | 524 | 99 / 75 |
| per-channel, attention kept fp32 | 68.7 [66.3-71.0] | 432 (28.92%) [26.67-31.27] | +6.33 [+3.78, +8.52] | 564 | 105 / 78 |

| Configuration | Accuracy [95% CI] | Decisions changed vs GPU fp32 | ΔECE (points) | MiB | ms/item (b1 / b16) |
|---|---|---|---|---|---|
| per-channel, FFN input only | 87.2 [85.4-88.8] | 28 (1.87%) [1.30-2.70] | +0.19 [-0.77, +0.89] | 635 | 110 / 99 |

## 6 Rounding weights to fewer bits

Weight-only rounding is where the 421 M parameters can lose the most bytes without touching the architecture. Table 6 and Figure 2 show the grid of bit width, scale granularity and scope.

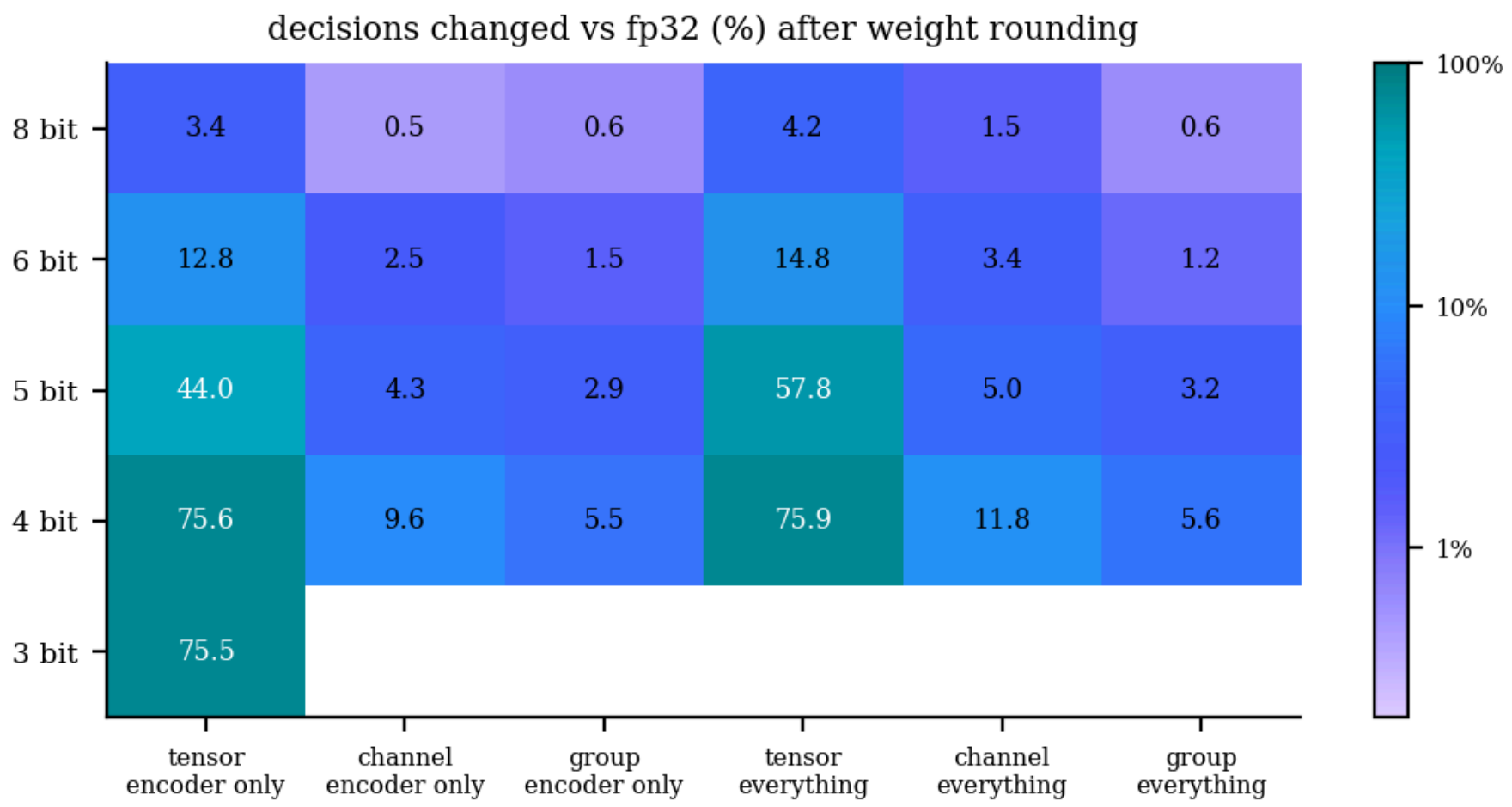

**Figure 2.** Decisions changed (%) against fp32 for every combination of bit width, scale granularity and scope (fine-tuned checkpoint). 'Everything' adds the token embeddings, option head, scorer and act head.

**The scale granularity decides everything at low bit width.** With one scale per tensor the fine-tuned model keeps its decisions to within 1% at none of 8/6/5/4/3 bits and within 5% at 8 bits; with one scale per output channel the thresholds are 8 bits and 5 bits, and with groups of 32 inputs 8 bits and 5 bits. Per-tensor 4-bit rounding changes 76% of all decisions and drops accuracy to 25%, below the 74% of the *untrained* checkpoint; per-channel 4-bit changes 9.6% and keeps accuracy at 85.2%. Grouping beats per-channel scales at 3 of the 4 bit widths we ran for both (8, 6, 5, 4 bits), at an extra 20 MiB for the additional scales at 4 bits. At 8 bits, per-channel rounding of the encoder changes 7 of 1,494 decisions; adding the embeddings, head, scorer and act head to the rounded set raises that to 22 (1.47% vs 0.47%).

**Table 6.** Weight rounding, fine-tuned checkpoint (and base checkpoint on the right). 'MiB' is the stored size with fp16 for everything not rounded and fp16 scales; fp32 reference accuracy is 87.3% (fine-tuned) and 73.6% (base).

| Configuration | MiB | Accuracy [95% CI] | Decisions changed | ΔECE (points) | Base: changed % | Base: accuracy % |
|---|---|---|---|---|---|---|
| 8-bit, channel scale, encoder | 477 | 87.5 [85.7-89.1] | 7 (0.47%) [0.23-0.96] | +0.13 [-0.67, +0.71] | 0.9 | 73.2 |
| 8-bit, channel scale, everything | 403 | 87.1 [85.4-88.8] | 22 (1.47%) [0.97-2.22] | +0.24 [-0.76, +1.16] | 2.0 | 73.5 |
| 8-bit, tensor scale, encoder | 476 | 86.1 [84.2-87.7] | 51 (3.41%) [2.61-4.46] | -0.03 [-1.38, +0.82] | 5.4 | 73.4 |
| 6-bit, channel scale, encoder | 395 | 86.7 [84.9-88.4] | 37 (2.48%) [1.80-3.39] | +0.30 [-0.83, +1.05] | 5.0 | 73.4 |
| 6-bit, group scale, encoder | 415 | 87.6 [85.8-89.1] | 23 (1.54%) [1.03-2.30] | +0.13 [-0.96, +0.89] | 3.3 | 73.2 |
| 6-bit, tensor scale, encoder | 394 | 82.0 [80.0-83.9] | 191 (12.78%) [11.19-14.57] | +2.31 [+0.63, +3.96] | 22.5 | 67.1 |

| Configuration | MiB | Accuracy [95% CI] | Decisions changed | ΔECE (points) | Base: changed % | Base: accuracy % |
|---|---|---|---|---|---|---|
| 5-bit, group scale, encoder | 374 | 86.8 [85.0-88.4] | 43 (2.88%) [2.14-3.85] | +0.43 [-0.69, +1.26] | 6.6 | 72.6 |
| 4-bit, channel scale, encoder | 313 | 85.2 [83.3-86.9] | 144 (9.64%) [8.24-11.24] | +0.46 [-1.43, +2.18] | 14.1 | 70.4 |
| 4-bit, group scale, encoder | 333 | 85.8 [83.9-87.5] | 82 (5.49%) [4.44-6.76] | -0.43 [-1.47, +0.65] | 10.6 | 71.6 |
| 4-bit, group scale, everything | 226 | 85.5 [83.7-87.2] | 84 (5.62%) [4.56-6.91] | -0.00 [-1.26, +0.88] | 10.5 | 71.2 |
| 4-bit, tensor scale, encoder | 313 | 25.3 [23.2-27.6] | 1129 (75.57%) [73.33-77.68] | -2.32 [-4.69, -0.36] | 73.2 | 23.3 |

**Which tensors carry the sensitivity.** Rounding a single group of tensors to 4 bits (per-channel) shows the ranking in Figure 3: attention QKV (6.8%), token embeddings (4.3%), FFN input (3.7%), FFN output (3.7%) are the most sensitive and option head (0.0%), scorer (0.0%), act head (0.0%) the least. At 3 bits the same ranking gives attention QKV (18%), token embeddings (10%), FFN input (13%) for the top three. Protecting the sensitive group in fp16 and rounding the rest is therefore an obvious lever, which the Pareto analysis in Section 19 takes up.

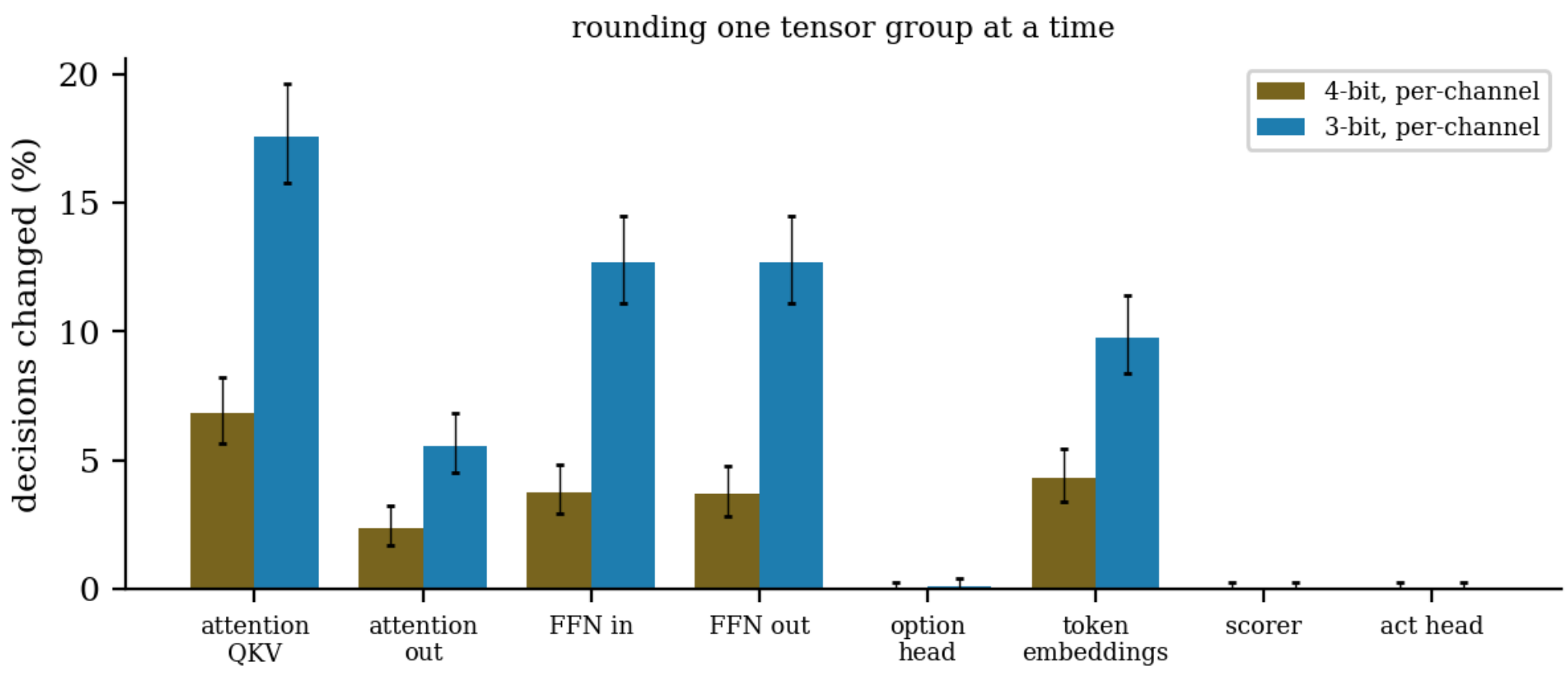

**Figure 3.** Decisions changed when only one tensor group is rounded (per-channel, 4 and 3 bits), fine-tuned checkpoint. Error bars are Wilson 95% intervals.

**Base checkpoint.** The untrained checkpoint shows the same pattern: per-tensor 4-bit rounding changes 73% of its decisions, per-channel 14.1% and grouped 10.6%. Because the base model is wrong on a quarter of the items anyway, its flip rate partly reflects instability on items it cannot decide; Section 10 separates the two.

## 7 Deleting encoder layers

Deleting layers shortens every forward pass in proportion. Table 7 and Figure 4 sweep top-*k*, bottom-*k*, middle blocks and regular patterns, always without retraining, and always from the fine-tuned and base checkpoints separately.

**The top of the stack is surprisingly expendable.** On the fine-tuned checkpoint, deleting the top 6 layers keeps the share of changed decisions at or below 1%, the top 7 at or below 2% and the top 9 at or below 5%; that corresponds to 348 M, 335 M and 311 M parameters left (of 421 M). The untrained checkpoint tolerates 4 (1%) and 5 (2%) top layers. Deleting the *bottom* layers is far worse: removing only the first layer changes 70% of decisions and leaves 30% accuracy. Removing every other layer (72% (odd layers) and 77% (even layers)) is not a way to halve the model without retraining.

**Comparison with FastJEV.** FastJEV [17] reports, in its abstract, that decision-guided pruning with no training cuts candidate depth by 43.75 to 45.83% and keeps 93.66 to 97.52% of the original task score on average. The closest points in our grid are the top 12 layers deleted (42.9% of the depth) and the top 13 (46.4%). At 12 layers our model keeps 66.6% of its fp32 accuracy and changes 40.6% of its decisions; at 13 layers it keeps 53.5% and changes 52.9%. These figures are not directly comparable with theirs: the model family, the tasks, the way depth is counted (candidate paths, not encoder layers) and the metric (task score, not agreement) all differ, and our top-*k* rule is position-based, not decision-guided. Our nearest equivalent of a decision-guided rule, the greedy search above, did worse than the top-*k* rule at every size we tried, so on this model we cannot reproduce the benefit of selecting layers by their effect on the answers.

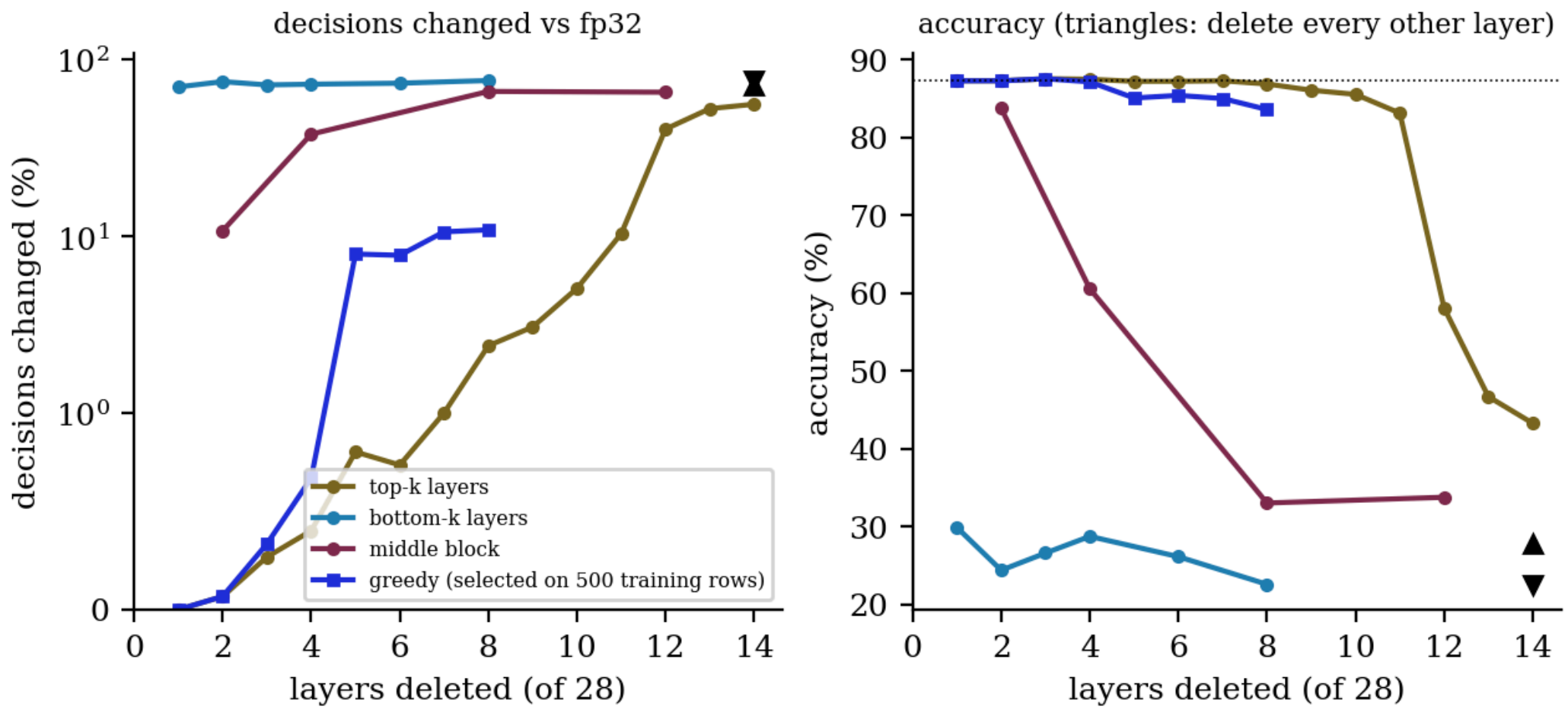

**Figure 4.** Decisions changed (left, symmetric-log axis) and accuracy (right) against the number of encoder layers deleted, by deletion pattern (fine-tuned checkpoint).

**Table 7.** Deleting encoder layers without any retraining (fine-tuned checkpoint). 'TV' is the mean shift of the answer probabilities in percent; the last column is the base checkpoint's share of changed decisions.

| Layers deleted | n | Params (M) | MiB | Accuracy [95% CI] | Decisions changed | ΔECE (points) | TV (%) | Base: changed % |
|---|---|---|---|---|---|---|---|---|
| top 2 | 2 | 397 | 757 | 87.3 [85.5-88.9] | 1 (0.07%) [0.01-0.38] | -0.52 [-1.42, +0.43] | 1.5 | 0.1 |
| top 4 | 4 | 372 | 710 | 87.5 [85.7-89.1] | 6 (0.40%) [0.18-0.87] | +0.30 [-2.13, +2.36] | 10.3 | 0.9 |
| top 6 | 6 | 348 | 663 | 87.2 [85.4-88.8] | 11 (0.74%) [0.41-1.31] | +33.33 [+30.18, +35.91] | 44.4 | 3.0 |
| top 8 | 8 | 323 | 617 | 86.9 [85.1-88.5] | 36 (2.41%) [1.75-3.32] | +49.71 [+46.24, +52.32] | 61.4 | 8.2 |
| top 10 | 10 | 299 | 570 | 85.5 [83.7-87.2] | 76 (5.09%) [4.08-6.32] | +53.70 [+50.33, +56.33] | 66.8 | 19.3 |
| top 14 | 14 | 250 | 476 | 43.3 [40.8-45.8] | 838 (56.09%) [53.56-58.59] | +14.30 [+10.74, +16.63] | 69.3 | 60.3 |
| bottom 1 | 1 | 409 | 780 | 29.9 [27.7-32.3] | 1050 (70.28%) [67.91-72.54] | +3.12 [+1.36, +5.78] | 65.8 | 65.8 |
| bottom 4 | 4 | 372 | 710 | 28.8 [26.5-31.1] | 1088 (72.82%) [70.51-75.02] | -0.81 [-2.80, +1.79] | 68.6 | 73.1 |
| middle 4 | 4 | 372 | 710 | 60.6 [58.1-63.0] | 567 (37.95%) [35.52-40.44] | +5.86 [+3.18, +8.36] | 47.3 | 48.0 |
| every other (odd) | 14 | 250 | 476 | 27.9 [25.7-30.2] | 1074 (71.89%) [69.55-74.11] | -1.06 [-3.36, +1.67] | 68.5 | 73.8 |
| every third | 9 | 311 | 593 | 61.8 [59.4-64.3] | 545 (36.48%) [34.08-38.95] | +7.14 [+4.82, +10.25] | 50.3 | 42.2 |

**Accuracy and decisions diverge.** The mean answer shift grows long before decisions flip: at the top 8 layers deleted the pooled TV is 61%, although only 2.4% of the decisions change. The top option survives, but the probability the model reports for it does not; Section 11 examines what that does to calibration, and it is the reason a deployment that thresholds on probability needs a recalibration step after layer deletion.

**One layer at a time.** A complete single-layer deletion sweep would need 28 more runs; only layers 0, 1, 2, 3 were run (changes of 70.3%, 13.3%, 9.2%, 7.5% of decisions), because the GPU was occupied by other jobs when the sweep was due and we did not start it. The per-layer picture therefore comes from rounding one layer at a time to 3 bits with every other layer in fp32 (Figure 5), which is a different intervention and a proxy for how much each layer matters. It changes between 0.0% and 7.5% of decisions. The four layers whose 3-bit rounding hurts most are 0, 1, 5, 7; the four least affected are 23, 24, 25, 26. The ten global-attention layers (0, 3, 6, ..., 27) average 3.0% against 2.9% for the eighteen windowed layers.

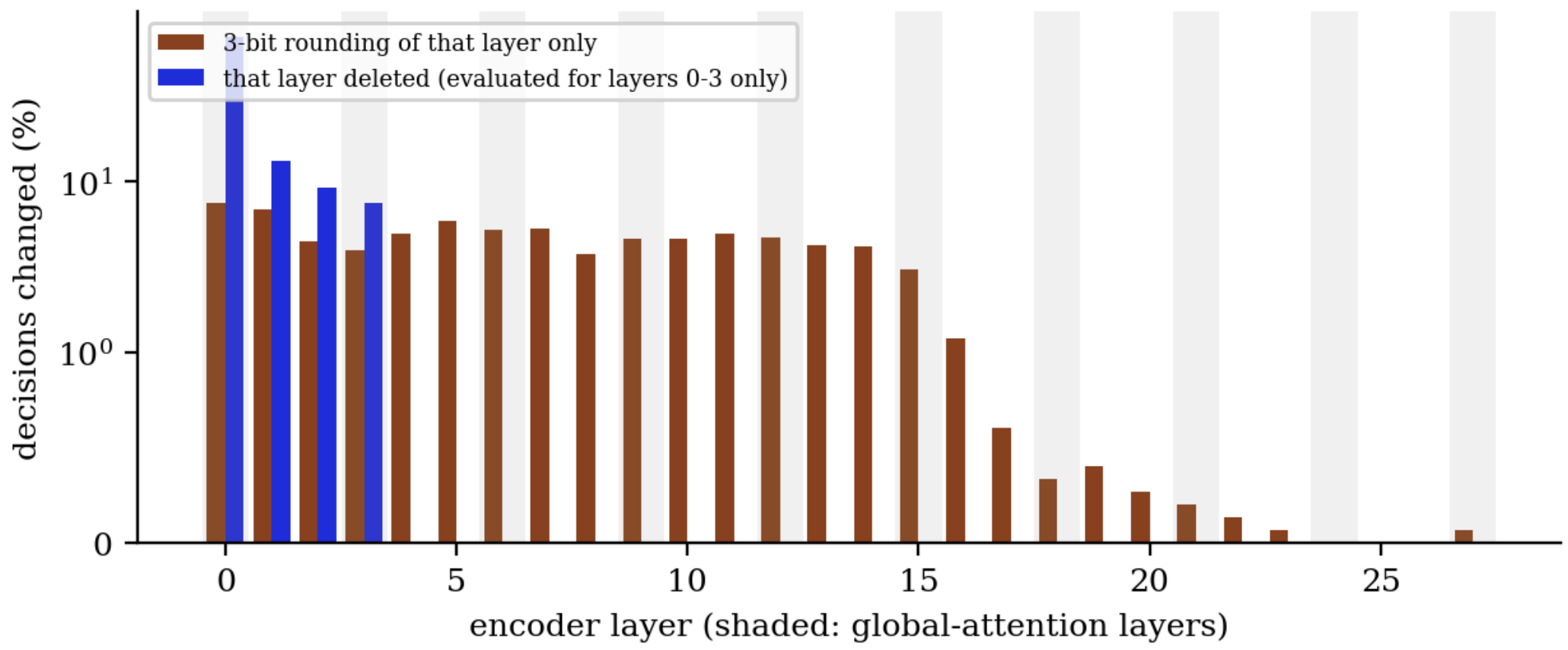

**Figure 5.** Per-layer sensitivity of the fine-tuned checkpoint: decisions changed when only that layer's weights are rounded to 3 bits (teal), and when that layer is deleted (orange; run for layers 0 to 3 only); symmetric-log axis; grey bands mark global-attention layers.

**Greedy selection.** A greedy search that deletes, at every step, the layer that moves the answer probabilities least on 240 training rows (no test item is used) finds sets that beat the top-k rule at 0 of 8 sizes. At 8 deleted layers it chooses layers 3, 13, 17, 20, 21, 24, 26, 27, with 10.9% flips versus 2.4% for the top eight.

## 8 Slicing and sparsity

Instead of deleting whole layers, one can thin every layer. We rank feed-forward units by the product of the norms of their gate, value and output weights, and attention heads by the norms of their value and output slices, keep the top fraction, and physically slice the matrices (so the thinner model is also faster). Table 8 gives the result without any retraining.

Slicing the feed-forward blocks to 90% changes 18.3%; 75% changes 35.5%; 50% changes 60.6%; 25% changes 74.2% of decisions; keeping 14 heads changes 11.0%; 12 heads changes 27.4%; 8 heads changes 56.4%; 4 heads changes 71.5%. Unstructured sparsity is the gentlest at moderate levels (10% zeros changes 2.8%; 20% zeros changes 5.3%; 30% zeros changes 7.6%; 50% zeros changes 18.8%; no level we tried stays within 1%), but a dense kernel does not skip zeros, so those points buy a smaller compressed file, not a faster forward pass. Taken alone none of the three structured knobs beats layer deletion on flips per parameter removed: compare the 335 M-parameter model with the top seven layers deleted (1.0% flips) with the FFN-0.5 model at 308 M (60.6%).

**Table 8.** Slicing and sparsity without retraining (fine-tuned checkpoint). 'Params' counts non-zero-shape parameters: sparsity leaves the dense shapes unchanged.

| Configuration | Params (M) | Accuracy [95% CI] | Decisions changed | ΔECE (points) | Base: changed % |
|---|---|---|---|---|---|

| Configuration | Params (M) | Accuracy [95% CI] | Decisions changed | ΔECE (points) | Base: changed % |
|---|---|---|---|---|---|
| keep 90% of FFN units | 399 | 78.2 [76.0-80.2] | 274 (18.34%) [16.46-20.38] | +9.83 [+7.92, +11.82] | n/a |
| keep 75% of FFN units | 365 | 64.0 [61.5-66.4] | 530 (35.48%) [33.09-37.94] | +9.98 [+7.77, +12.55] | n/a |
| keep 50% of FFN units | 308 | 39.2 [36.8-41.7] | 905 (60.58%) [58.07-63.02] | -1.00 [-2.75, +2.06] | n/a |
| keep 25% of FFN units | 252 | 25.4 [23.2-27.6] | 1108 (74.16%) [71.88-76.32] | -2.93 [-4.69, -0.40] | n/a |
| keep 14 of 16 heads | 407 | 85.1 [83.2-86.8] | 164 (10.98%) [9.49-12.66] | +0.75 [-0.84, +2.34] | n/a |
| keep 12 of 16 heads | 392 | 73.7 [71.4-75.9] | 409 (27.38%) [25.18-29.69] | +8.05 [+4.94, +10.32] | n/a |
| keep 8 of 16 heads | 363 | 44.0 [41.5-46.6] | 843 (56.43%) [53.90-58.92] | +6.06 [+3.07, +8.42] | n/a |
| keep 4 of 16 heads | 333 | 27.3 [25.1-29.6] | 1068 (71.49%) [69.14-73.72] | -2.15 [-4.39, +0.50] | n/a |
| zero 10% of weights | 421 | 87.0 [85.2-88.6] | 42 (2.81%) [2.09-3.78] | +0.32 [-0.83, +1.30] | n/a |
| zero 30% of weights | 421 | 85.2 [83.3-86.9] | 114 (7.63%) [6.39-9.09] | -0.14 [-1.72, +1.21] | n/a |
| zero 50% of weights | 421 | 77.4 [75.3-79.5] | 281 (18.81%) [16.91-20.87] | +4.72 [+1.94, +6.90] | n/a |
| zero 90% of weights | 421 | 22.9 [20.8-25.1] | 1164 (77.91%) [75.74-79.94] | -4.84 [-6.54, -2.59] | n/a |

## 9 Trimming the vocabulary

The embedding table is 12.2% of the parameters and, unlike the encoder, is a pure lookup: removing rows costs no compute, only memory, and changes a decision only if a removed token occurs in an input. The 4,379 training texts (all four training pools) use 12,999 of the 50,368 token ids; the 1,494 test items use 8,661, of which 1,314 never occur in training; 1.1% of test tokens are such out-of-vocabulary tokens. Figure 6 shows the frequency distribution and the trade-off.

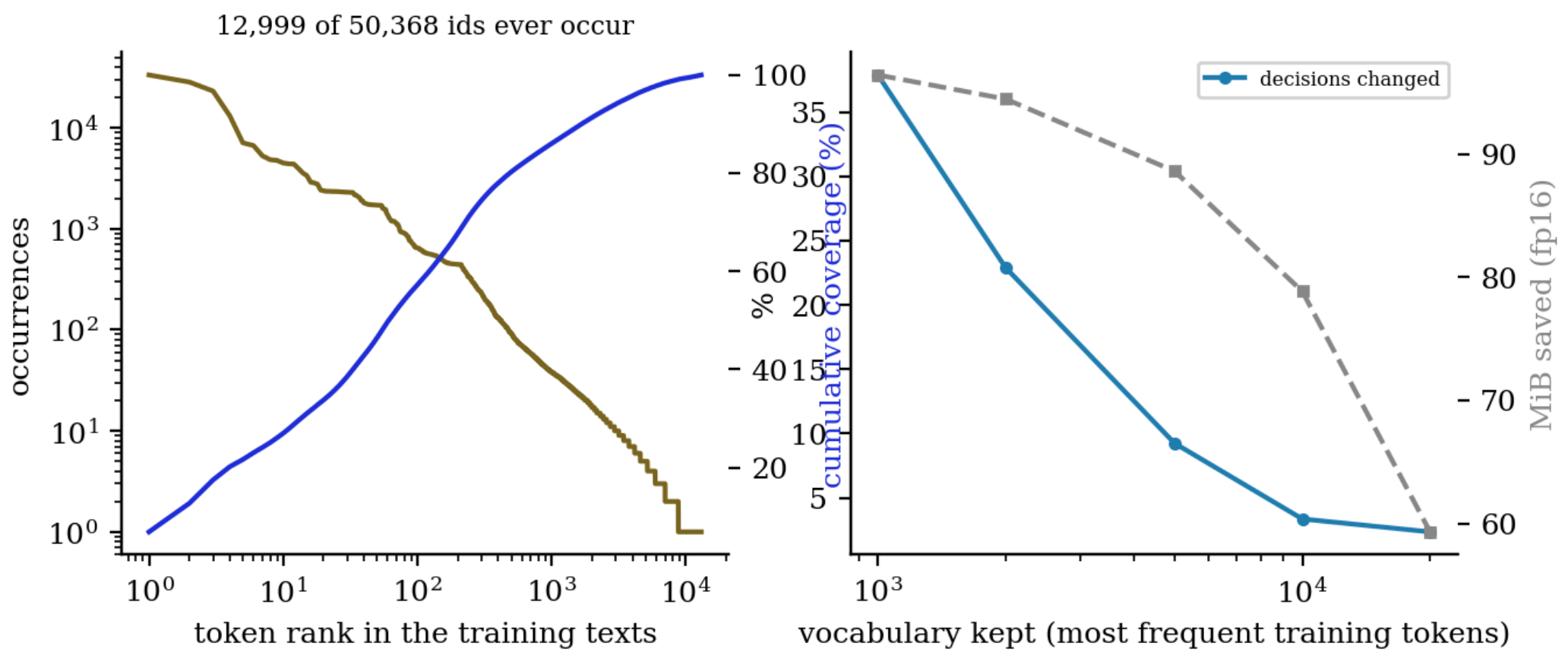

**Figure 6.** Token frequency in the training texts (left; orange: cumulative share of occurrences) and, for the most-frequent-N vocabularies, decisions changed and memory saved (right). Fine-tuned checkpoint.

**Table 9.** Vocabulary trimming (fine-tuned checkpoint): rows of the 50,368-row embedding table not kept are replaced by the [UNK] row (or zeros). 'Saved' is the fp16 memory of the removed rows.

| Configuration | Rows kept | Saved (MiB) | Accuracy [95% CI] | Decisions changed |
|---|---|---|---|---|
| keep training + test ids, zero the rest | 14,398 | 70 | 87.3 [85.5-88.9] | 0 (0.00%) [0.00-0.26] |
| keep training ids, rest to [UNK] | 13,084 | 73 | 86.3 [84.4-87.9] | 35 (2.34%) [1.69-3.24] |

| Configuration | Rows kept | Saved (MiB) | Accuracy [95% CI] | Decisions changed |
|---|---|---|---|---|
| keep 20,000 most frequent | 13,084 | 73 | 86.3 [84.4-87.9] | 35 (2.34%) [1.69-3.24] |
| keep 10,000 most frequent | 10,085 | 79 | 85.5 [83.7-87.2] | 50 (3.35%) [2.55-4.39] |
| keep 5,000 most frequent | 5,085 | 88 | 80.3 [78.2-82.3] | 138 (9.24%) [7.87-10.81] |
| keep 2,000 most frequent | 2,085 | 94 | 69.2 [66.8-71.5] | 342 (22.89%) [20.83-25.09] |
| keep 1,000 most frequent | 1,085 | 96 | 55.2 [52.6-57.7] | 566 (37.88%) [35.46-40.37] |

Keeping every id that appears in training or test leaves every decision unchanged (0 changes), as it must, and releases 70 MiB; this is the check that the trimming code touches nothing it should not, not a deployable setting because it peeks at the test ids. The deployable version keeps the ids seen in training and sends the rest to [UNK]: it changes 35 of 1,494 decisions (2.34% [1.69-3.24]) and frees 73 MiB. Keeping only the most frequent 5,000 ids frees 89 MiB and changes 9.2% [7.9-10.8] of decisions. The cost of trimming is therefore a distribution-shift cost: a new language or a new domain brings tokens the trimmed table has dropped, and the multilingual and code-mixed sets are where we would expect the damage; Figure 7 gives the per-set rates for the 5,000-id vocabulary. With only 12,999 ids in the training texts, a 20,000-id cutoff keeps the same rows as the training ids.

## 10 Which decisions move first

Pooled flip rates hide who pays. Figure 7 breaks the fine-tuned checkpoint's flips down by test set for ten representative recipes, and Figure 8 groups items by question type: few-option choice (up to five options), many-option choice (six or more), ordinal score questions and yes/no questions.

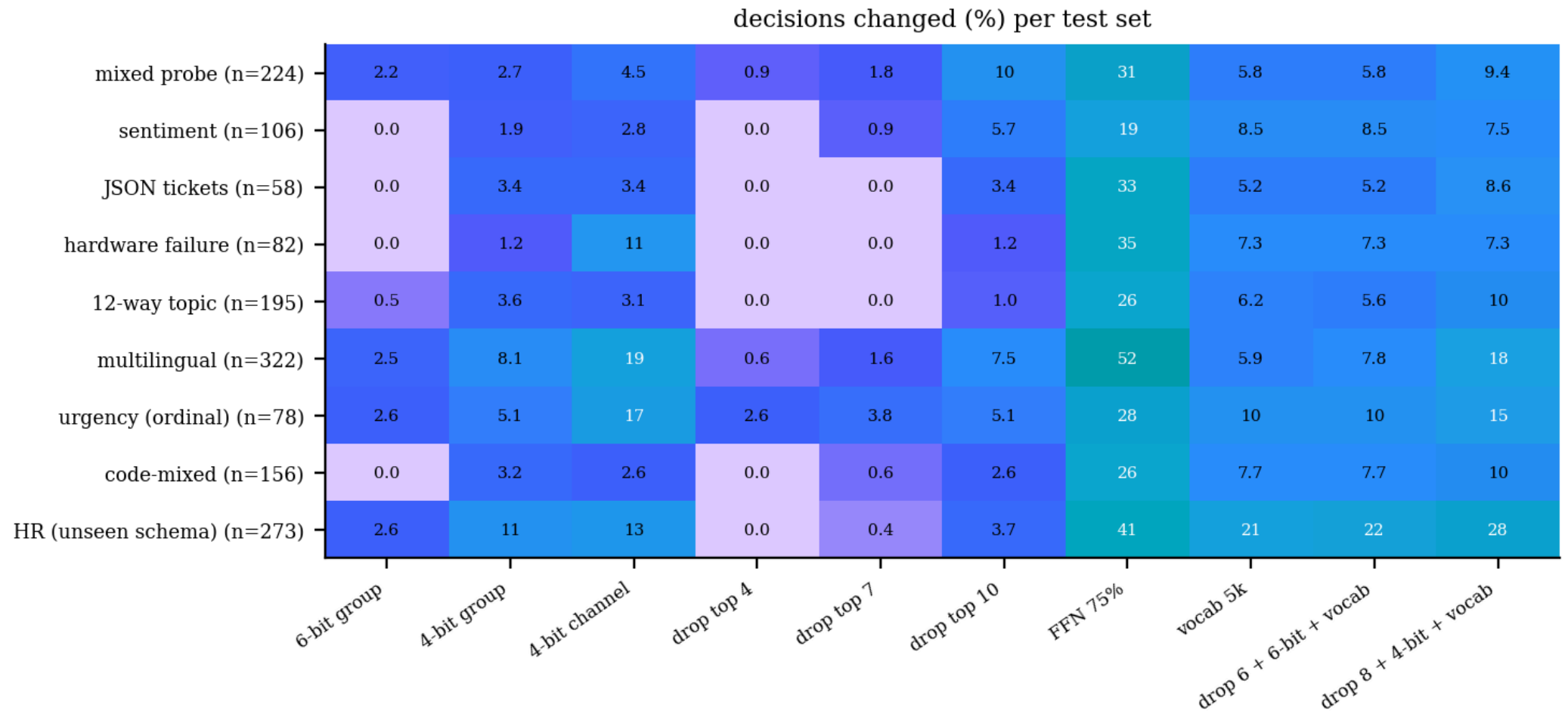

**Figure 7.** Decisions changed (%) per test set and recipe, fine-tuned checkpoint. Each cell is a rate over the n items of that set (n in the row labels), so small sets such as the 58 JSON items are noisy.

Across the ten recipes the set with the highest flip rate is HR (unseen schema) for 4 of them and urgency (ordinal) for 3. By question type, the most fragile group is 'choice K>=6' in 4 of the 9 recipes that change at least 8 decisions. The comparisons that matter for a deployment decision are paired and have intervals:

- **Many-option questions (six or more options, 543 items)** against all others: 4-bit group +2.1 points [-0.3, +4.8]; drop top 7 -0.7 points [-1.6, +0.2]; 4-bit channel -2.1 points [-5.1, +1.2].
- **Ordinal score questions (89 items)**: 4-bit group -1.1 points [-4.9, +3.8]; drop top 7 +2.5 points [-0.8, +6.9]; 4-bit channel +5.3 points [-1.5, +13.5].
- **Multilingual and code-mixed items (478 items)**: 4-bit group +1.5 points [-1.2, +4.2]; drop top 7 +0.4 points [-0.7, +1.5]; 4-bit channel +6.1 points [+2.5, +9.6].

Intervals that include zero mean the data do not separate that group from the rest at this sample size; we read the signs but not the sizes of the others. The ordinal set has only 78 items, the three-level scale leaves little room between neighbouring levels, and that set is the smallest, so its estimates are the least reliable of the four question types.

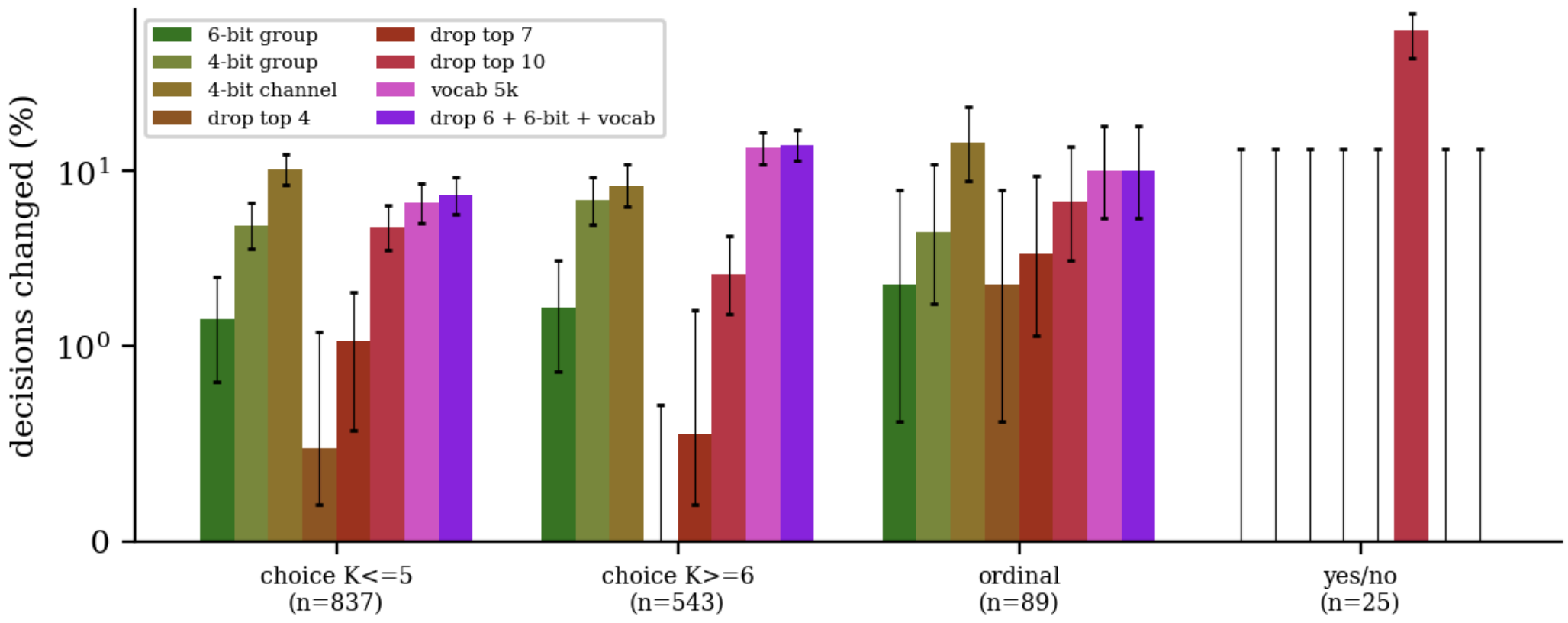

**Figure 8.** Decisions changed by question type for eight recipes (fine-tuned checkpoint); symmetric-log axis.

### 10.1 Do the same recipes break the untrained checkpoint more?

Figure 9 applies the same recipes to both checkpoints. The base checkpoint changes more decisions than the fine-tuned one for 9 of the 9 recipes shown. Averaged over the recipes, the flip rate is 1.73% on items the fine-tuned fp32 model gets right and 11.46% on items it gets wrong; for the base checkpoint the corresponding figures are 4.58% and 13.00%.

> **Reading this** Part of what looks like fragility of the base model is that a quarter of its decisions sit on items it cannot decide (its accuracy is 74% against 87% for the fine-tuned model): when the top two options are nearly tied, any perturbation reverses them. Fine-tuning widens the gaps, and the compressed models inherit that.

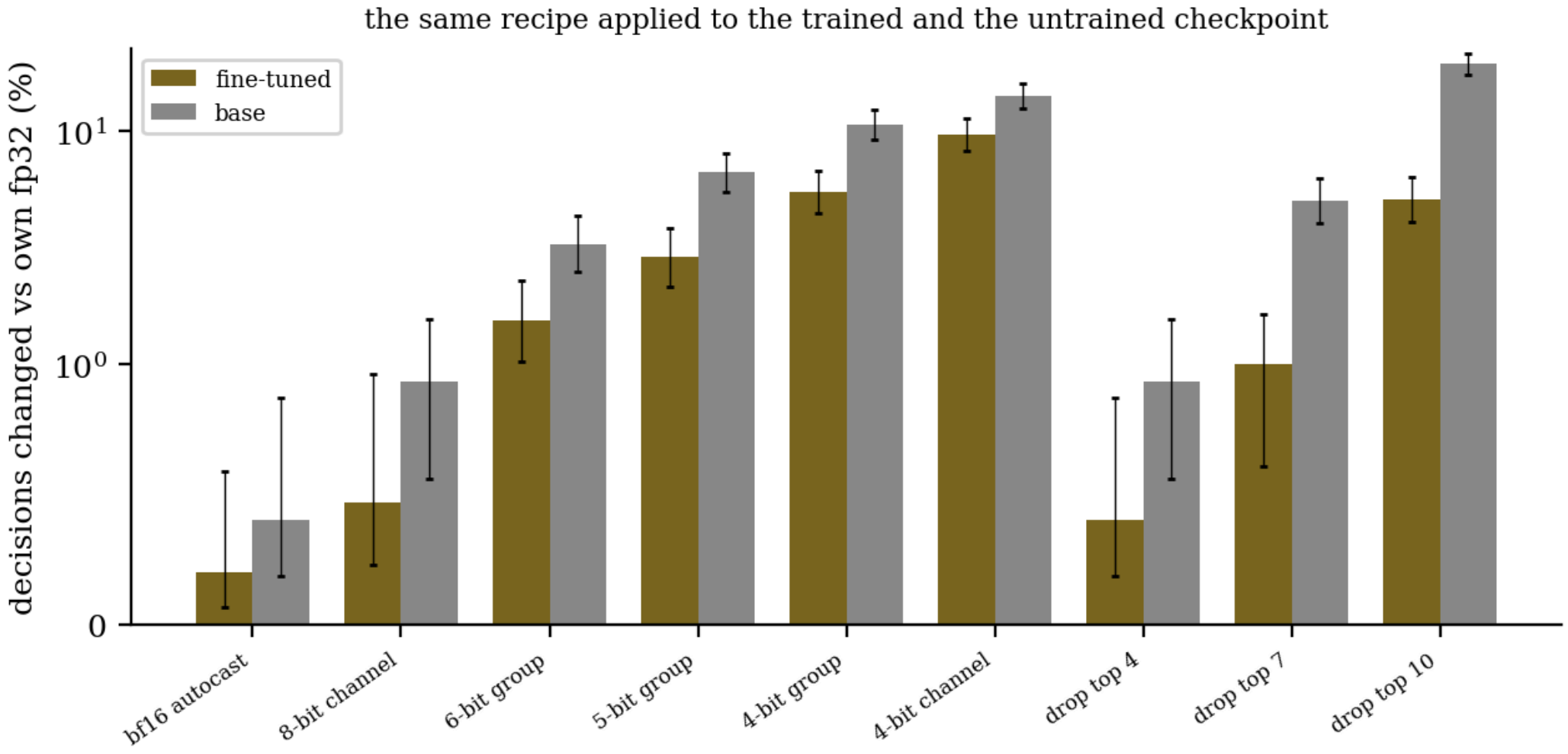

**Figure 9.** Share of decisions changed by the same recipe on the fine-tuned and on the untrained checkpoint (each against its own fp32). Wilson 95% intervals; symmetric-log axis. Recipes not run on the base checkpoint are left out.

## 11 Calibration under compression

An application that thresholds the reported probability (to abstain, escalate or route) cares about the *value* of the top probability, not only whether the top option is right. Figure 10 and Table 10 show how ECE moves. The reference fp32 ECE is 5.80%. 7 of the 13 recipes shown move it by an amount whose paired bootstrap interval excludes zero; the largest move is +53.7 points for 'drop top 10'. Recipes that lower mean confidence by more than two points: 4-bit channel, drop top 4, drop top 7, drop top 10, greedy 8 layers, FFN 75%, vocab 5k, drop 6 + 6-bit + vocab, drop 8 + 4-bit + vocab.

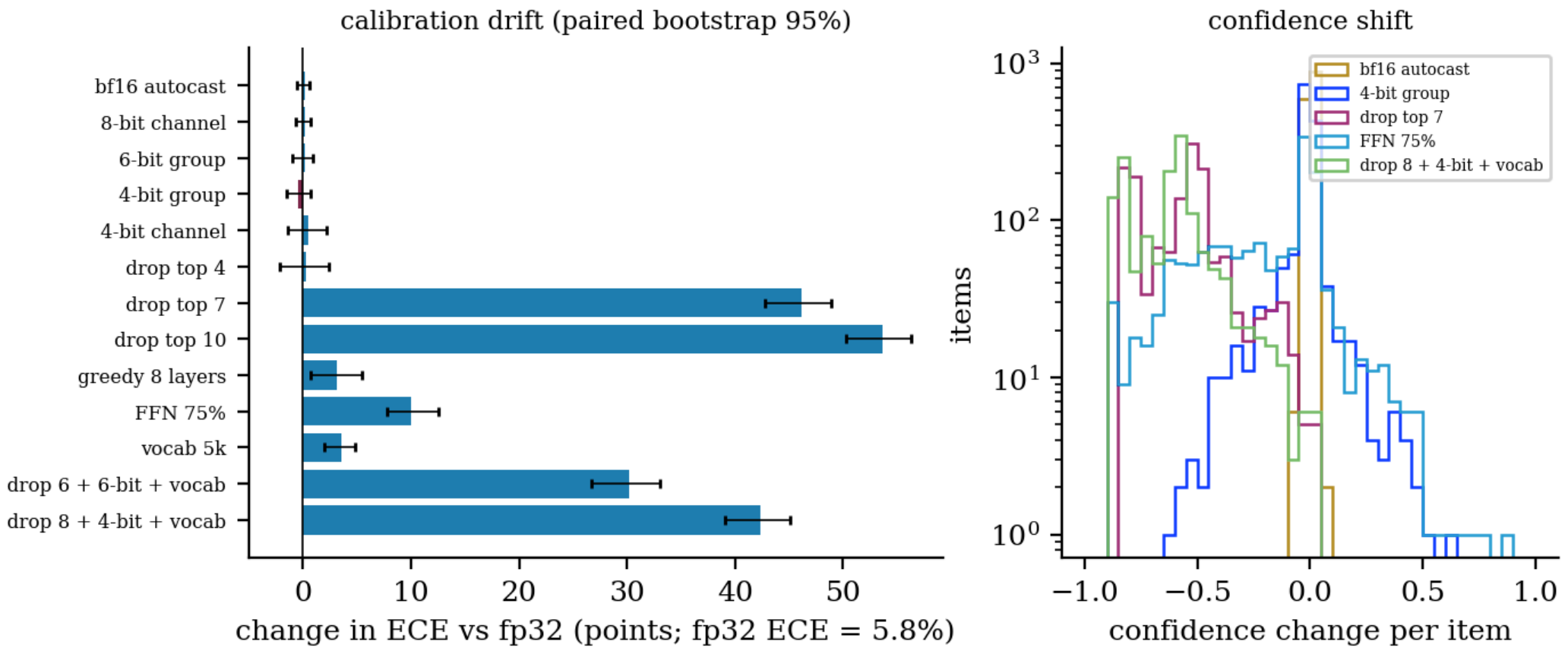

**Figure 10.** Change in ECE against fp32 (left; fp32 ECE 5.8%) and the per-item change in top-option confidence for four recipes (right).

**Table 10.** Calibration under compression, fine-tuned checkpoint. ECE is the 10-bin expected calibration error on all 1,494 items (fp32: 5.80%; mean confidence 92.3%; accuracy 87.3%). 'Recal.' is the ECE after a cross-fitted isotonic map fitted on four fifths of the items and applied to the rest; for fp32 itself it is 2.17%.

| Recipe | ECE % | ΔECE (points) [95% CI] | Mean conf. % | TV % | Recal. ECE % |
|---|---|---|---|---|---|
| bf16 autocast | 5.93 | +0.13 [-0.57, +0.63] | 92.2 | 0.3 | 3.05 |
| 8-bit channel | 5.93 | +0.13 [-0.67, +0.71] | 92.2 | 0.6 | 2.55 |
| 6-bit group | 5.93 | +0.13 [-0.96, +0.89] | 92.4 | 1.6 | 3.54 |
| 4-bit group | 5.37 | -0.43 [-1.47, +0.65] | 90.4 | 6.2 | 2.13 |
| 4-bit channel | 6.27 | +0.46 [-1.43, +2.18] | 87.5 | 12.1 | 3.30 |
| drop top 4 | 6.10 | +0.30 [-2.13, +2.36] | 82.3 | 10.3 | 1.87 |
| drop top 7 | 52.00 | +46.19 [+42.76, +48.92] | 35.3 | 57.4 | 0.94 |
| drop top 10 | 59.50 | +53.70 [+50.33, +56.33] | 26.1 | 66.8 | 1.55 |
| greedy 8 layers | 8.93 | +3.13 [+0.72, +5.43] | 74.7 | 21.3 | 1.33 |
| FFN 75% | 15.78 | +9.98 [+7.77, +12.55] | 72.2 | 40.6 | 4.63 |
| vocab 5k | 9.41 | +3.60 [+2.00, +4.80] | 88.5 | 10.8 | 3.16 |
| drop 6 + 6-bit + vocab | 36.05 | +30.24 [+26.67, +33.06] | 44.8 | 48.7 | 2.79 |
| drop 8 + 4-bit + vocab | 48.15 | +42.35 [+39.08, +45.10] | 29.7 | 63.4 | 1.71 |

**Is the drift recoverable?** The last column recalibrates each compressed model with an isotonic map (cross-fitted, so no item calibrates itself). If that column is close to the fp32 value of 2.17% the drift is a monotone distortion of confidence that a recalibration pass on a few hundred labelled items could remove; where it stays above, the compression has changed *which* items the model is sure about. For the layer-deletion recipes the recalibrated ECE is 0.94% to 1.55% (fp32: 2.17%) although their raw ECE is 52.0% to 59.5%: after the top layers are deleted the model still ranks its items almost as well as before, but the numbers it reports are on a different scale (mean confidence falls from 92% to 35% at seven layers). A deployment that thresholds on probability would need to refit its thresholds or a calibration map after deleting layers. A single refitted temperature does most of that repair, with or without labels (Section 16). Recalibration was done on test items by cross-fitting and is therefore indicative: a deployment would fit it on held-in data.

### 11.1 Where do the flips live?

Figure 11 splits flips by the fp32 model's own confidence and by whether fp32 was right. For the six recipes shown, the flip rate in the lowest-confidence quintile is 23.4% on average against 2.62% in the highest. A deployment that already routes low-confidence answers to a person therefore exposes few of the flipped decisions to users, which is a practical reason to measure flips *conditional on confidence* and not only in aggregate.

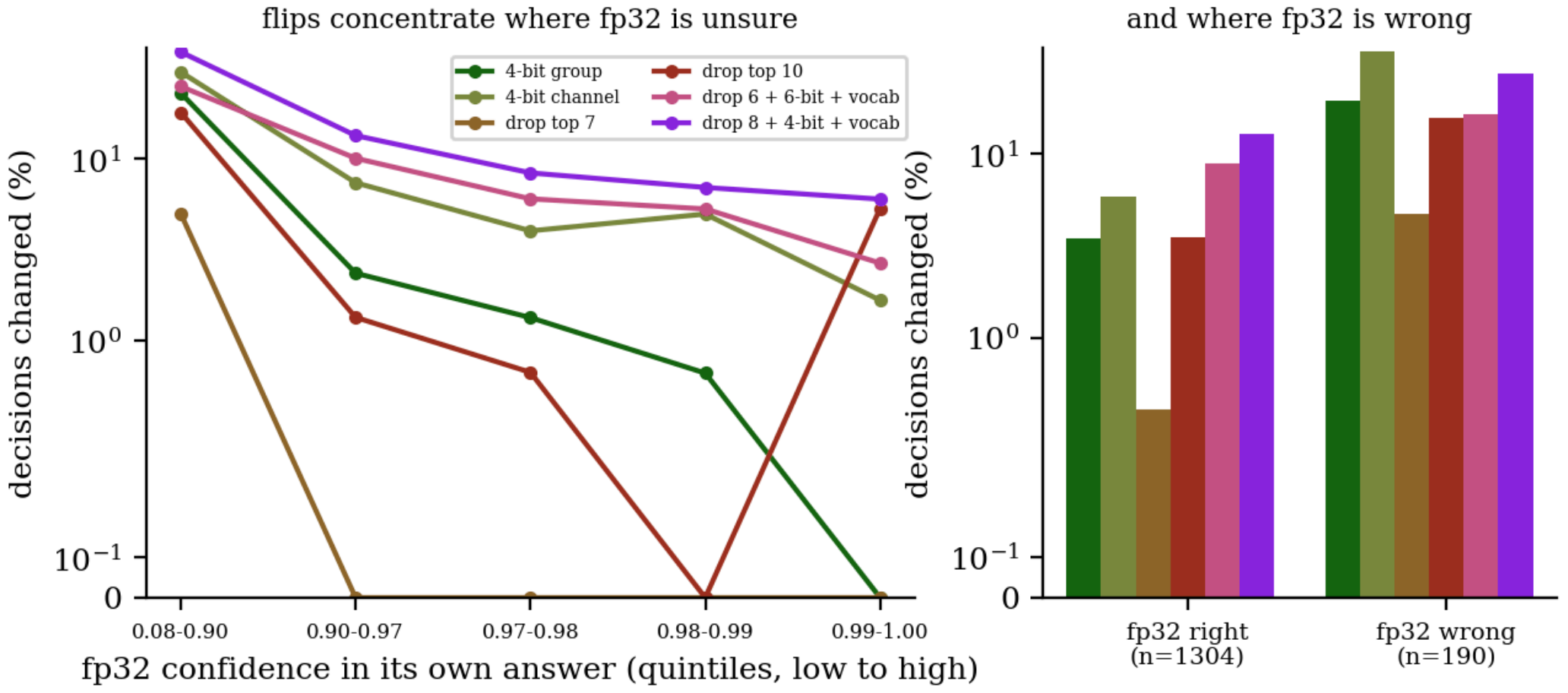

**Figure 11.** Decisions changed by the fp32 model's confidence quintile (left) and by whether fp32 was right (right); fine-tuned checkpoint, symmetric-log axes.

## 12 Held-out schemas and public benchmarks

A study of this kind has to say what the evaluation items share with the model's training. The fine-tuned checkpoint was trained on 2,818 synthetic cases written with 6 option schemas (a schema is a question type plus its named options, for example billing, technical or other). Eight of the nine synthetic test sets use exactly those schemas with new items: the mixed probe, sentiment, JSON tickets, hardware failures, the 12-way topic set, urgency, and the multilingual and code-mixed sets (the last two change the language of the text but not the schema). We call these eight sets *in-family*. The ninth, HR, has a 12-way schema with option names that appear nowhere in training, so we call it the *unseen-schema* set (273 items). Every headline number in the first part of this report pools all nine. To address the obvious objection that these are synthetic items from the same generator, we add 800 questions from four public benchmarks, ARC-Easy, ARC-Challenge, MMLU and HellaSwag, 200 each, drawn with a fixed seed from the benchmarks' test or validation files. Each becomes one row: the question (or the HellaSwag context) is the state, the options are named A to D and their descriptions are the answer texts. Nothing from these benchmarks was seen in training, the option names, the topic, the text style and the task all differ, and the model was never told that the descriptions are answers. We call them *out of family*. They are a stress test of the decision-flip metric, not of benchmark accuracy: the model scores 37% on them, which is close to chance for some of the four (see Table 11).

**Table 11.** The fp32 fine-tuned model on the three item groups. 'Margin' is the log-probability gap between its top option and the runner-up (median); the last column is the share of items with a gap under 1 (top option less than 2.7 times as likely as the runner-up). Chance accuracy is 25% on the public sets (four options) and 8% on the 12-way HR set.

| Item group | Items | Accuracy | Mean confidence | ECE | Median margin | Margin < 1 |
|---|---|---|---|---|---|---|
| in-family (8 synthetic sets) | 1221 | 88.0% | 94.1% | 6.4% | 5.7 | 4% |
| unseen schema (HR) | 273 | 84.2% | 84.2% | 7.5% | 4.9 | 7% |
| public benchmarks (out of family) | 800 | 36.8% | 56.8% | 20.0% | 0.7 | 59% |

Because the model is unsure on these rows (median margin 0.7 against 5.7 in family; 59% of public-benchmark items have a margin under 1 against 4% in family), a small change to its scores can reorder the top two options. That is what we see. Across the 35 configurations whose in-family flip rate is above zero, the public-benchmark rate is at least as high as the in-family rate in 35 of them, and the median ratio is 4.0 among configurations that change at least 0.5% of in-family decisions (6-bit grouped rounding changes 1.3% in family and 5.9% on the public sets; 4-bit grouped rounding changes 4.3% in family and 21.4% on the public sets; deleting 6 layers changes 0.9% in family and 9.1% on the public sets). The unseen-schema set is at least as high as the in-family rate in 17 of 35 configurations, so a new schema moves flips less consistently than leaving the family altogether.

**Table 12.** Decisions changed against the model's own fp32 answer, by item group (fine-tuned checkpoint; count, rate, Wilson 95% interval). The last column divides the public-benchmark rate by the in-family rate.

| Recipe | MiB | In-family (1,221) | Unseen schema (273) | Public benchmarks (800) | Ratio |
|---|---|---|---|---|---|
| bf16 autocast | 804 | 3 (0.2%) [0.1-0.7] | 0 (0.0%) [0.0-1.4] | 18 (2.2%) [1.4-3.5] | 9.2 |
| fp16 autocast | 804 | 1 (0.1%) [0.0-0.5] | 0 (0.0%) [0.0-1.4] | 1 (0.1%) [0.0-0.7] | 1.5 |
| 8-bit, per-channel | 477 | 6 (0.5%) [0.2-1.1] | 1 (0.4%) [0.1-2.0] | 27 (3.4%) [2.3-4.9] | 6.9 |
| 6-bit, groups of 32 | 415 | 16 (1.3%) [0.8-2.1] | 7 (2.6%) [1.2-5.2] | 47 (5.9%) [4.4-7.7] | 4.5 |
| 4-bit, groups of 32 | 333 | 53 (4.3%) [3.3-5.6] | 29 (10.6%) [7.5-14.8] | 171 (21.4%) [18.7-24.3] | 4.9 |
| 4-bit, per-channel | 313 | 109 (8.9%) [7.5-10.7] | 35 (12.8%) [9.4-17.3] | 292 (36.5%) [33.2-39.9] | 4.1 |
| drop top 4 layers | 710 | 6 (0.5%) [0.2-1.1] | 0 (0.0%) [0.0-1.4] | 30 (3.8%) [2.6-5.3] | 7.6 |
| drop top 6 layers | 663 | 11 (0.9%) [0.5-1.6] | 0 (0.0%) [0.0-1.4] | 73 (9.1%) [7.3-11.3] | 10.1 |
| drop top 8 layers | 617 | 30 (2.5%) [1.7-3.5] | 6 (2.2%) [1.0-4.7] | 194 (24.2%) [21.4-27.3] | 9.9 |
| keep 90% of FFN units | 761 | 213 (17.4%) [15.4-19.7] | 61 (22.3%) [17.8-27.7] | 412 (51.5%) [48.0-54.9] | 3.0 |
| keep 12 of 16 heads | 748 | 337 (27.6%) [25.2-30.2] | 72 (26.4%) [21.5-31.9] | 454 (56.8%) [53.3-60.1] | 2.1 |
| zero 30% of weights | 804 | 90 (7.4%) [6.0-9.0] | 24 (8.8%) [6.0-12.7] | 235 (29.4%) [26.3-32.6] | 4.0 |
| drop 4 + 8-bit | 406 | 13 (1.1%) [0.6-1.8] | 1 (0.4%) [0.1-2.0] | 44 (5.5%) [4.1-7.3] | 5.2 |
| drop 6 + 8-bit | 383 | 12 (1.0%) [0.6-1.7] | 1 (0.4%) [0.1-2.0] | 85 (10.6%) [8.7-13.0] | 10.8 |

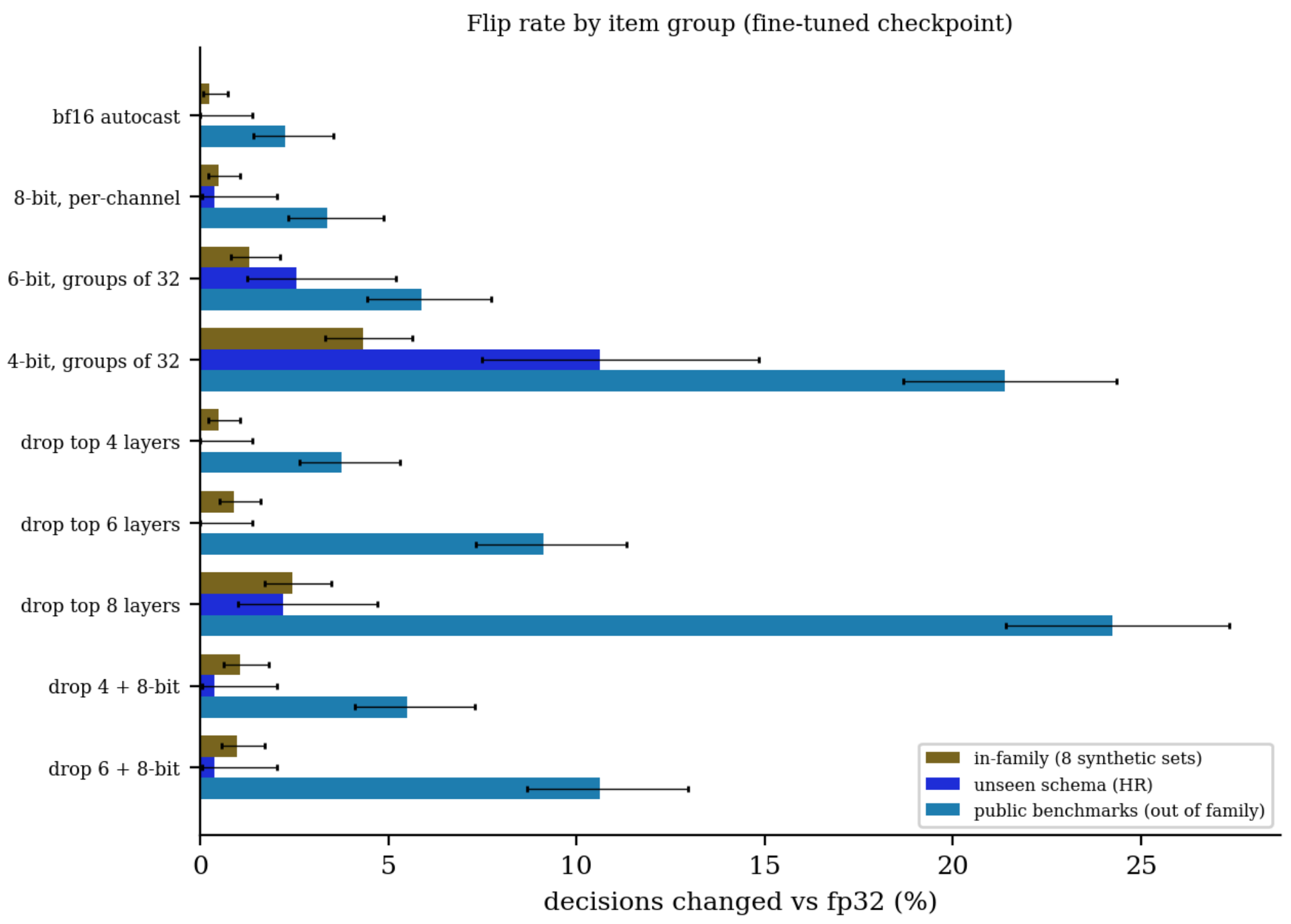

**Figure 12.** Decisions changed against fp32 in each item group for nine recipes (fine-tuned checkpoint, Wilson 95% intervals). The public-benchmark bars are the out-of-family result.

**What this changes.** A recipe that looks free in family is not always free out of family. bf16 changes 0.2% of in-family decisions and 2.2% of public-benchmark decisions; 8-bit per-channel rounding changes 0.5% and 3.4%; the combination 'drop 4 + 8-bit', which Section 19 files under tier A on the nine synthetic sets, changes 1.1% in family and 5.5% out of family. The recipes of tier A therefore carry the label 'in-family' and should be rechecked on items that look like the deployment, which are not ours. Not every flip on these rows is a loss: of the 171 decisions that 4-bit grouped rounding changes there, fp32 was right and the rounded model wrong on 52, the reverse on 48; for deleting eight layers the counts are 53 and 57 of 194. Accuracy of the recipes in Table 12 on these rows stays within 8 points of fp32. The point of the group is that the *decision* is unstable where the model is unsure, which is also where a user would be least able to tell.

**Table 13.** Decisions changed (%) on each public benchmark (200 rows each), fine-tuned checkpoint.

| Recipe | ARC-Easy | ARC-Challenge | MMLU | HellaSwag |
|---|---|---|---|---|
| bf16 | 1.0 | 3.0 | 3.5 | 1.5 |
| 6-bit group | 2.0 | 9.5 | 5.0 | 7.0 |
| 4-bit group | 18.5 | 22.5 | 24.0 | 20.5 |
| drop 6 | 5.0 | 9.0 | 11.0 | 11.5 |
| drop 8 | 15.0 | 26.5 | 21.5 | 34.0 |

## 13 Noise floor and other checkpoints

A comparison of two builds changes more than the weights, and Blagojevich and Sunderland [22] measure how much for two decision models: engine, restart and batch differences moved up to 0.92% of decisions in their setting. We run the same check for Laya on one GPU with one engine. Re-running fp32 with batch sizes of 1, 4 and 64, and with the rows in reverse order, changes none of the 2,294 decisions and at most 0.0001 of probability mass on any item (Table 14). Every recipe that changes at least one decision is therefore above the floor: bf16 autocast (21), fp16 autocast (2), drop top 2 layers (13), 8-bit, per-channel (34), 6-bit, groups of 32 (70), 4-bit, groups of 32 (253). The floor is small here because the runs share one device, one library version and one set of kernels, so it is a lower bound on what a change of runtime would add.

**Table 14.** Same fp32 weights, different batching or row order, against the standard fp32 run (batch 16, file order), on all 2,294 items. Nothing was compressed in these runs.

| Run | Decisions changed | Share [95% CI] | Largest probability change (TV) |
|---|---|---|---|
| fp32 rerun, batch 16 | 0 | 0.00% [0.00-0.17] | 0.0000 |
| fp32, batch 1 | 0 | 0.00% [0.00-0.17] | 0.0001 |
| fp32, batch 4 | 0 | 0.00% [0.00-0.17] | 0.0001 |
| fp32, batch 64 | 0 | 0.00% [0.00-0.17] | 0.0001 |
| fp32, rows reversed, batch 16 | 0 | 0.00% [0.00-0.17] | 0.0001 |
| fp32, rows reversed, batch 1 | 0 | 0.00% [0.00-0.17] | 0.0001 |

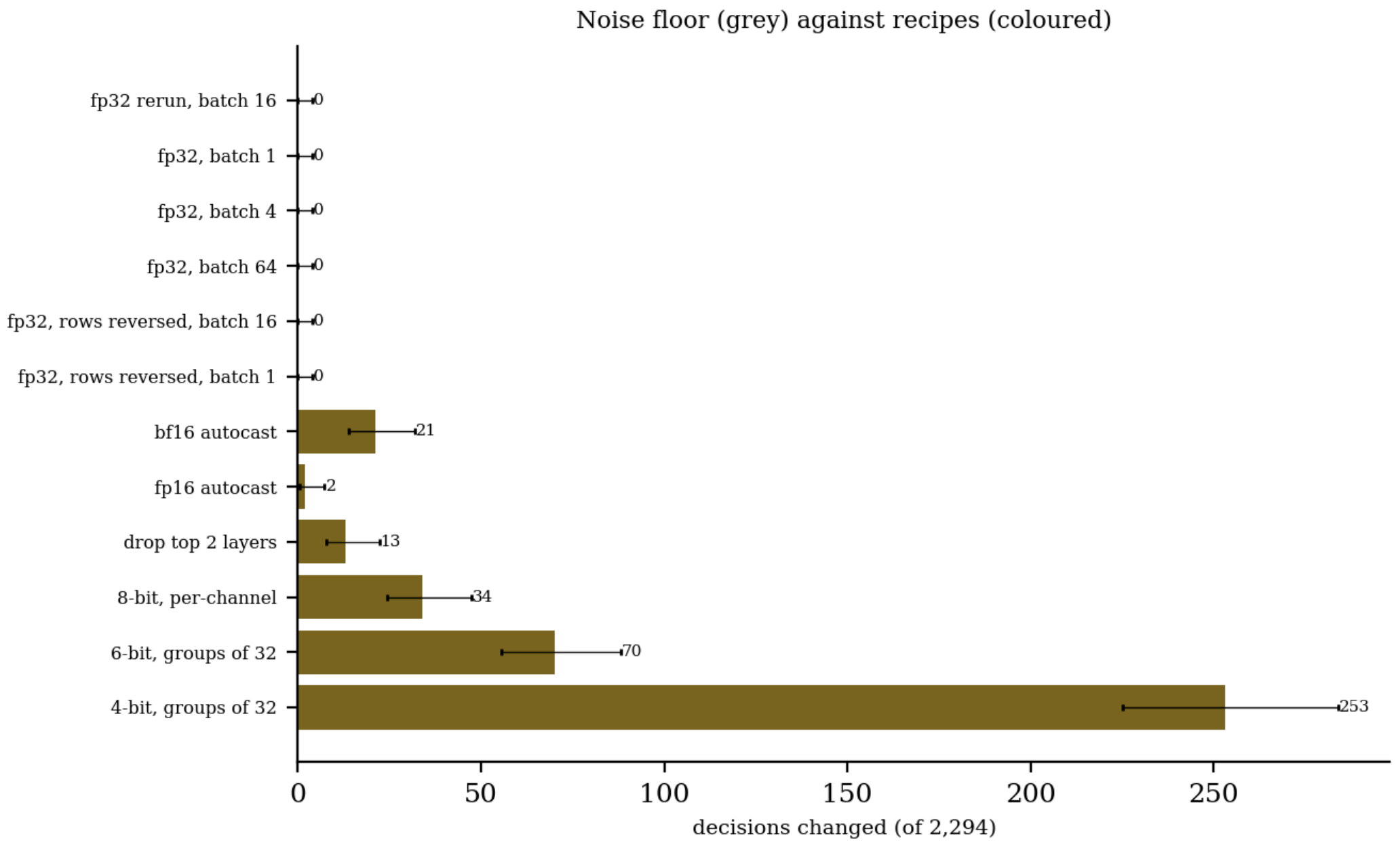

**Figure 13.** Noise floor (grey bars) against recipes (coloured bars): decisions changed out of 2,294 items, with Wilson intervals (fine-tuned checkpoint).

**Reproduction.** The second-round sweep re-ran 35 configurations of the first-round sweep on the same 1,494 synthetic items, with the stored fp32 answers as reference in both. The flip counts differ by at most 0 items, and 0 of the 35 configurations differ at all, so the first-round numbers reproduce on this machine and the noise-floor runs measure the effect of batching, not of rerunning.

### 13.1 Is it one checkpoint?

Every number so far is for one fine-tuned checkpoint. Table 15 repeats ten recipes on 4 more: base (untrained), fine-tune seed 1, fine-tune seed 2, other backbone (mmBERT-base). All were trained or published with the same trainer. The two seeds use the same training rows as the main checkpoint. The other backbone is a Laya-style model built on mmBERT-base (22 layers of width 768, 322M parameters) and trained on the multilingual rows, so 'drop top k' there means the top k of its own layers. The base model has no task training. Among the 3 fine-tunes on the main training rows, the standard deviation of the flip rate is 0.2 points for 6-bit grouped rounding, 0.7 for 4-bit grouped and 0.1 for dropping six layers. The ordering of the ten recipes by flip rate on the main checkpoint matches that of the other checkpoints with Kendall's $\tau$ of base (untrained) 1.00, fine-tune seed 1 1.00, fine-tune seed 2 1.00, other backbone (mmBERT-base) 0.78. Rounding behaves alike on the other backbone (4-bit grouped: 11.0% on the main checkpoint and 13.2% there; 6-bit grouped: 3.1% and 3.6%), but layer deletion does not: deleting six layers changes 3.7% on the main checkpoint and 21.4% there, and FFN slicing 45.3% and 62.4%. A recipe card therefore transfers across retrains of the same recipe and, for rounding, across backbones; the depth that can be deleted is a property of the model and has to be measured again.

**Table 15.** The same ten recipes on every checkpoint we evaluated (decisions changed against that checkpoint's own fp32, % of all 2,294 items; the second column is its fp32 accuracy on the nine synthetic sets).

| Checkpoint | fp32 acc. % | bf16 | 8-bit channel | 6-bit group | 4-bit group | 4-bit tensor | drop 4 | drop 6 | drop 10 | FFN 75% | drop 6 + 8-bit |
|---|---|---|---|---|---|---|---|---|---|---|---|
| fine-tuned (v1) | 87.3 | 0.9 | 1.5 | 3.1 | 11.0 | 74.9 | 1.6 | 3.7 | 15.3 | 45.3 | 4.3 |
| base (untrained) | 73.6 | 0.7 | 1.7 | 4.8 | 15.0 | 72.8 | 2.1 | 6.5 | 28.0 | 54.4 | 7.3 |
| fine-tune seed 1 | 87.2 | 0.7 | 1.3 | 3.6 | 12.6 | 74.1 | 1.9 | 3.6 | 16.5 | 49.1 | 4.1 |
| fine-tune seed 2 | 87.1 | 0.7 | 1.4 | 3.4 | 11.6 | 73.8 | 1.8 | 3.5 | 15.1 | 45.6 | 4.1 |
| other backbone (mmBERT-base) | 85.1 | 0.5 | 1.4 | 3.6 | 13.2 | 67.1 | 6.3 | 21.4 | 69.2 | 62.4 | 21.9 |

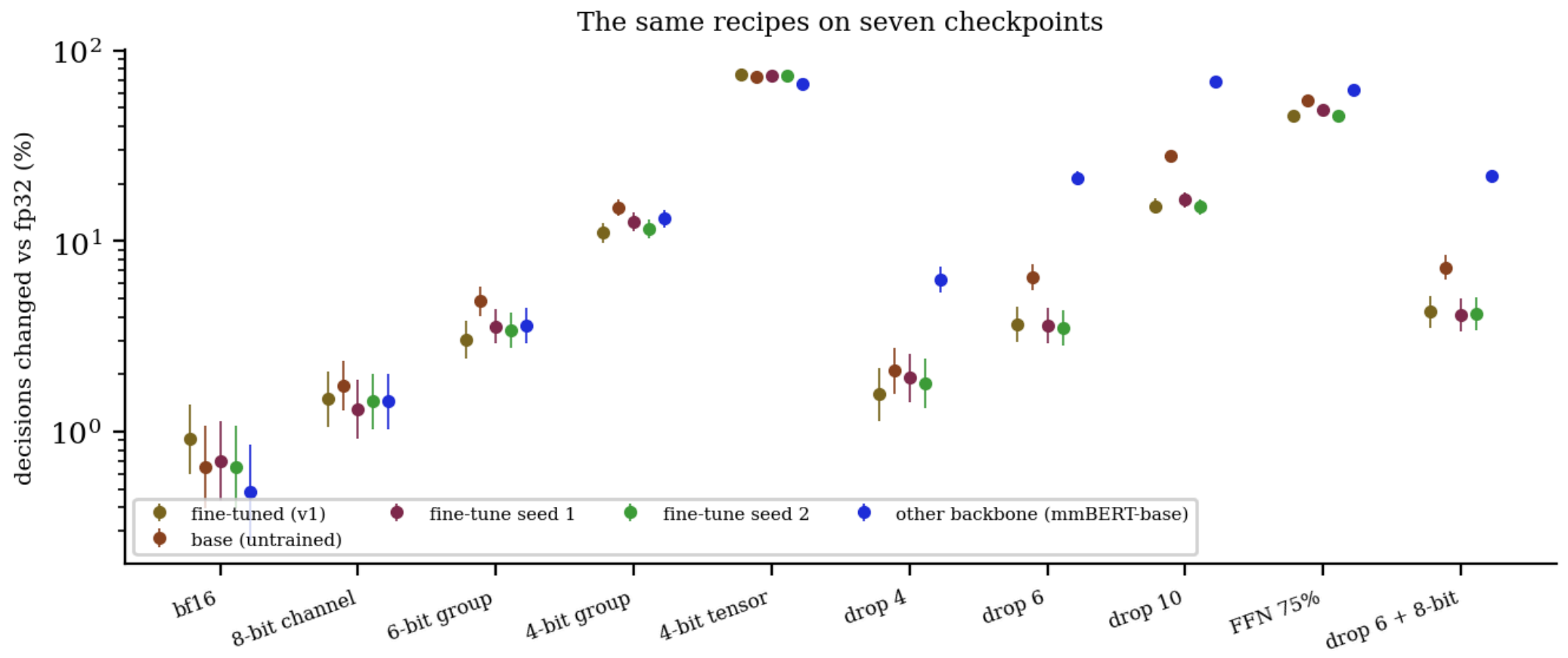

**Figure 14.** The ten recipes on every checkpoint we evaluated: decisions changed against that checkpoint's own fp32 (all items, Wilson 95% intervals, log axis).

## 14 Why decisions flip: total variation and margins

The earlier sections count flips. This one asks what predicts them, using two accounts from the literature. Both use the probabilities we stored for every item and configuration, and both were stated as predictions before we looked (the list is in Section 1).

### 14.1 Total variation

Almeida Felicio [20] reports, for compressed language models, that the rate at which the arg-max token changes is close to the mean total variation between the dense and the compressed distribution (median ratio 1.05 over 802 copies of 19 models). A one-pass decision encoder has a closed set of options, which makes the same check cheap. Table 16 gives the ratio of our flip rate to the mean TV by knob family and item group. For weight rounding in family the ratio has median 0.92 (range 0.81 to 1.08), so one unit of answer shift costs close to one flip, as that paper found for language models. Layer deletion breaks the relation: the median in-family ratio is 0.07, because deleting layers moves the reported probabilities a long way (a shift of scale) without changing most arg-max decisions, the effect Section 11 describes. On the public-benchmark rows the ratio is above one for rounding (1.83) and for slicing and sparsity (1.81), against our pre-stated expectation that it would fall below 0.6 when confidence is already low: the expectation was wrong. A likely reason is the margins of Table 11: when many items sit near a tie, a given amount of probability moved is more likely to land on an item that flips. We did not test that separately. Over all configurations with at least 0.5% flips the ratio lies between 0.5 and 2 for 69% of configurations in family, 79% on the unseen schema and 57% out of family. TV is a useful guide to flips for rounding, and a poor one for deletion.

**Table 16.** Flip rate divided by mean total variation (TV) between the compressed and the fp32 probabilities, median [range] over configurations whose flip rate in that group is at least 0.5%. A ratio of 1 means one flipped decision per unit of TV.

| Knob family | in-family (8 synthetic sets) | unseen schema (HR) | public benchmarks (out of family) |
|---|---|---|---|
| weight rounding | 0.92 [0.81-1.08] (n=9) | 0.87 [0.64-1.23] (n=9) | 1.83 [1.63-2.50] (n=10) |
| layer deletion | 0.07 [0.02-0.83] (n=10) | 0.07 [0.02-0.74] (n=7) | 0.59 [0.15-2.03] (n=14) |
| FFN, head and sparsity | 0.87 [0.65-0.94] (n=7) | 0.88 [0.52-1.06] (n=7) | 1.81 [1.61-2.14] (n=7) |
| combinations | 0.12 [0.02-0.74] (n=3) | 0.75 [0.75-0.75] (n=1) | 0.36 [0.27-1.96] (n=3) |

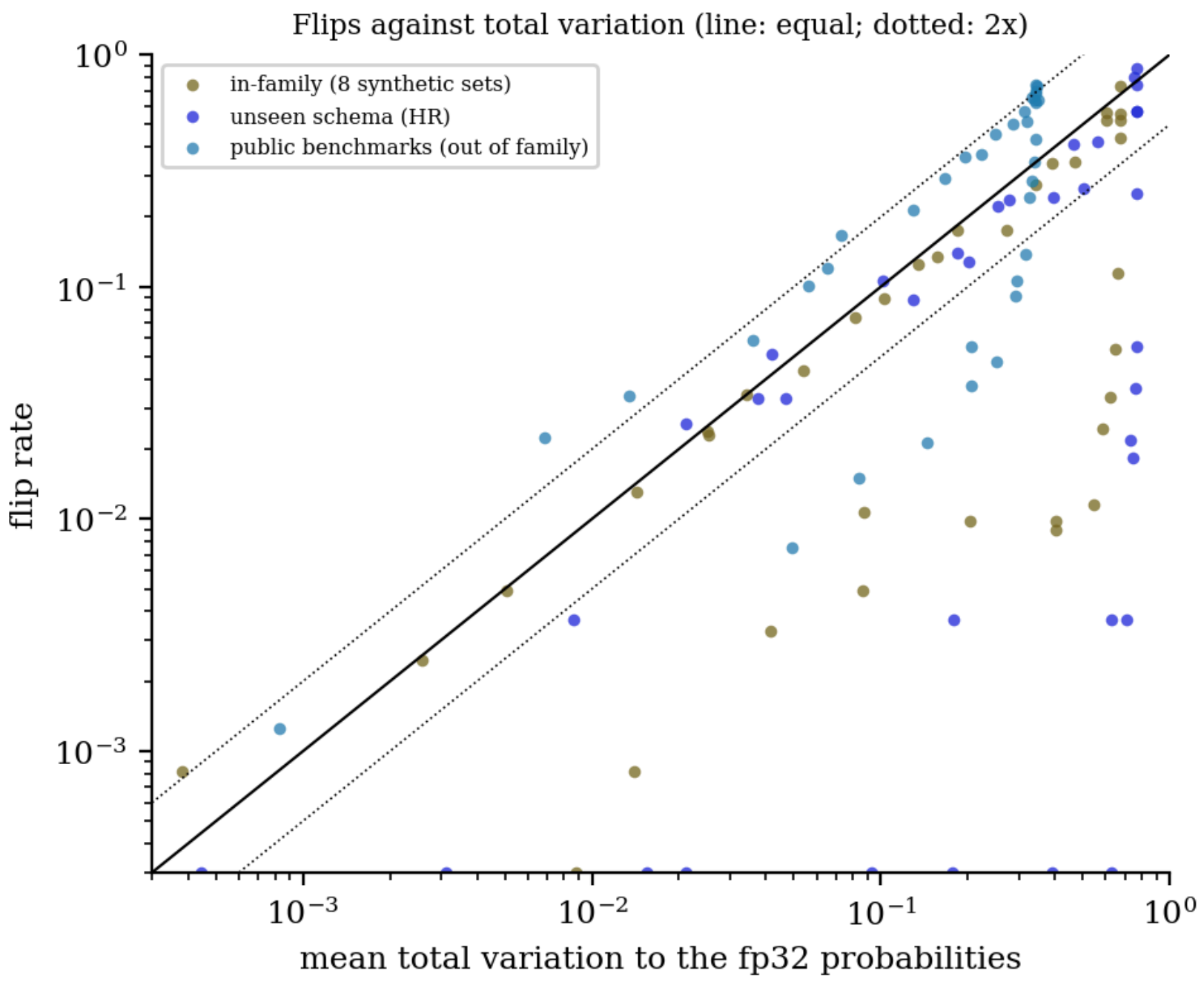

**Figure 15.** Mean total variation against flip rate for every configuration in each item group (log axes). The line is one flip per unit of TV; dotted lines are a factor of two.

## 14.2 Margins

Wu et al. [21] propose that quantization shrinks decision margins roughly linearly, m′ ≈ c·m + b with c usually below one, and that the fit predicts flip rates for 16 language models. Here, for weight rounding, the margin after rounding is close to the margin before plus a constant: the fitted slope lies between 0.95 and 1.01, the offset moves from -0.02 at 8 bits to -1.23 at 3 bits, and $R^2$ falls from 1.00 to 0.39. So for this encoder the damage of rounding is a downward shift with growing noise, not a shrinkage towards zero. The fit predicts the flip rate of held-out items to within 1.3 points at every bit width (Table 17). For the structural knobs it does not: the median error is up to 9.0 points and $R^2$ down to 0.26, because deleting a layer or slicing a block probably rescales the scores in a way that a single line with a common residual cannot describe. The simpler statement that survives for every knob is that flips live at small margins: the fp32 margin alone ranks the flipped items above the unflipped ones with an AUROC of 0.72 to 0.99.

**Table 17.** A linear account of the margin. The margin of an item is the log-probability of the fp32 model's top option minus that of the best other option; after compression it is measured for the same option, so a negative margin is a flip. We fit margin_after = slope × margin_before + offset on a random half of the 1,494 synthetic items and predict the flip rate of the other half; 'error' is the median absolute error in points over 20 random splits. The last column is the AUROC with which the fp32 margin alone ranks flipped items above unflipped ones.

| Recipe | Slope | Offset | Residual SD | $R^2$ | Observed flips | Predicted flips | Error (pts) | AUROC of fp32 margin |
|---|---|---|---|---|---|---|---|---|
| 8-bit, per-channel | 1.00 | -0.02 | 0.14 | 1.00 | 0.5% | 0.6% | 0.1 | 0.99 |
| 6-bit, groups of 32 | 0.99 | +0.04 | 0.43 | 0.96 | 1.5% | 1.1% | 0.5 | 0.98 |
| 5-bit, groups of 32 | 1.01 | -0.10 | 0.74 | 0.89 | 2.9% | 1.9% | 0.8 | 0.97 |
| 4-bit, groups of 32 | 1.01 | -0.43 | 1.30 | 0.72 | 5.4% | 4.2% | 1.1 | 0.91 |
| 3-bit, groups of 32 | 0.95 | -1.23 | 2.47 | 0.39 | 15.3% | 13.7% | 1.3 | 0.78 |
| drop top 4 layers | 1.00 | -1.31 | 0.94 | 0.83 | 0.4% | 7.6% | 7.3 | 0.98 |
| drop top 8 layers | 0.11 | -0.18 | 0.21 | 0.55 | 2.4% | 11.6% | 9.0 | 0.91 |
| keep 90% of FFN units | 1.00 | -1.26 | 3.43 | 0.27 | 18.3% | 17.5% | 1.6 | 0.78 |
| keep 12 of 16 heads | 0.65 | -1.61 | 2.27 | 0.26 | 27.4% | 26.4% | 1.5 | 0.72 |

| Recipe | Slope | Offset | Residual SD | $R^2$ | Observed flips | Predicted flips | Error (pts) | AUROC of fp32 margin |
|---|---|---|---|---|---|---|---|---|
| zero 30% of weights | 0.97 | -0.61 | 1.80 | 0.55 | 7.6% | 6.6% | 1.1 | 0.86 |

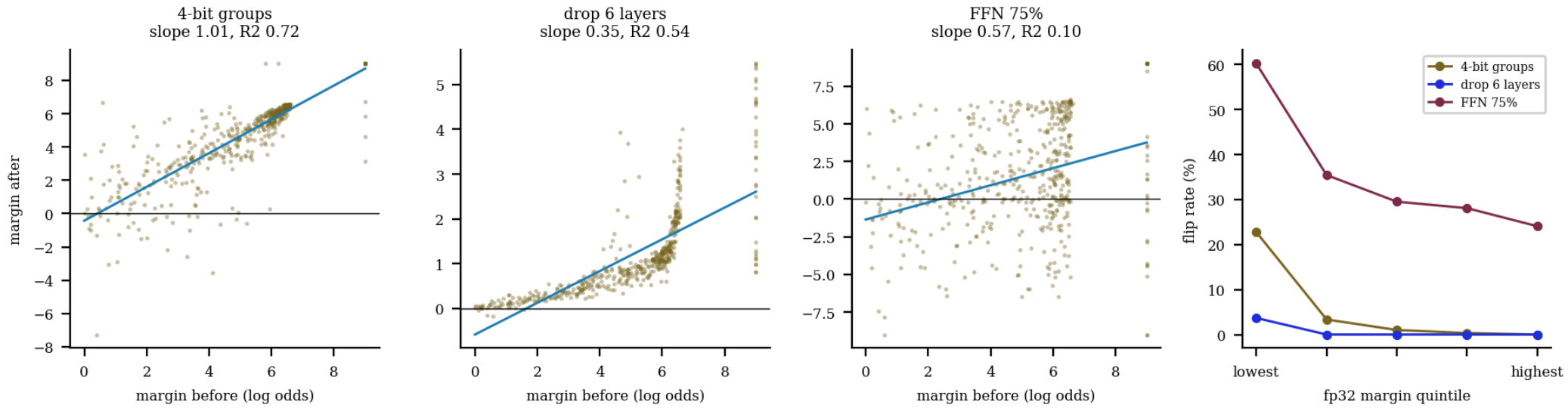

**Figure 16.** Margin after compression against margin before (500 random synthetic items, clipped at 9; solid line: least-squares fit), and flip rate by quintile of the fp32 margin (right).

**Use.** The margin is available at no cost from the fp32 model on any input, so a deployment that must audit a compressed model can compute it for a sample of its own traffic and read off how many items sit near a tie. The share of items with a margin below one is 4% in family and 59% on the public sets, and that difference alone accounts for much of why the same recipe flips more decisions out of family.

## 15 Early exit instead of deletion

Deleting layers statically costs the same on every input. A model that can stop early spends less on easy inputs and more on hard ones, as in confident adaptive decoding [27] and confidence cascades [28]. Our case is simple: the option head can be attached after any encoder layer, and the compressed models of Section 7 are exactly those early exits (the head was never trained for them). We test the natural rule. At each exit compute the model's top probability and stop if it reaches a threshold; otherwise carry on from the same hidden states to the next exit, down to the full 28 layers. The cost is the number of layers run. Because deleted-layer models are exact prefixes, nothing is recomputed. We stated before the run that a gated exit would use fewer layers than the best static deletion at the same flip budget.

**Table 18.** Early exit with the fp32 model's own head at shallower depths. An item leaves at the first exit whose top-option probability reaches the threshold; otherwise it runs all 28 layers. The threshold is chosen on the in-family items as the one that gives the fewest mean layers at 1% or fewer flips, then applied unchanged to the unseen-schema and public-benchmark items. 'Layers' is the mean number of encoder layers run; head cost is ignored. 'Gate recalibrated' rescales each exit's probabilities with one temperature fitted on labelled in-family items before thresholding.

| Method | Threshold | In-family layers | flips | Unseen layers | flips | Public layers | flips |
|---|---|---|---|---|---|---|---|
| static: delete top 2 | n/a | 26 | 0.1% | 26 | 0.0% | 26 | 1.5% |
| static: delete top 4 | n/a | 24 | 0.5% | 24 | 0.0% | 24 | 3.8% |
| static: delete top 6 | n/a | 22 | 0.9% | 22 | 0.0% | 22 | 9.1% |
| static: delete top 8 | n/a | 20 | 2.5% | 20 | 2.2% | 20 | 24.2% |
| early exit: one exit at 20 layers | 0.36 | 23.0 | 0.7% | 28.0 | 0.0% | 28.0 | 0.0% |
| early exit: one exit at 20 layers, gate recalibrated | 0.50 | 20.4 | 0.8% | 21.3 | 0.0% | 26.3 | 0.0% |
| early exit: exits at 16, 20, 24 layers | 0.53 | 24.1 | 0.9% | 25.5 | 0.0% | 27.5 | 0.0% |
| early exit: exits at 16, 20, 24 layers, gate recalibrated | 1.00 | 27.4 | 0.8% | 28.0 | 0.0% | 28.0 | 0.0% |
| early exit: exits at every depth from 14 to 27 | 0.55 | 22.8 | 0.7% | 24.5 | 0.0% | 27.0 | 0.0% |
| early exit: exits at every depth from 14 to 27, gate recalibrated | 1.00 | 26.4 | 0.8% | 27.7 | 0.0% | 28.0 | 0.0% |

The raw gate fails the prediction. The best cascade with exits at every depth runs 22.8 layers on average at 0.7% flips in family, against 22 layers and 0.9% for deleting 6 layers outright. The reason shows in the threshold's fate out of family: the mean top probability at the 20-layer exit is 36% in family and 27% on the public sets, so a threshold that is right in family is almost never reached there, and the cascade sends nearly every unseen-schema item (mean 24.5 layers) and public item (27.0 layers) to the full model. That is safe (flips 0.0% and 0.0%) and saves little.

Rescaling the exit probabilities first changes the result. With one temperature fitted on labelled in-family items for each exit (Section 16), a single exit at 20 layers runs 20.4 layers on average at 0.8% flips in family, 21.3 layers at 0.0% on the unseen schema and 26.3 layers at 0.0% on the public sets. In family that is 1.6 layers fewer than static deletion at the same flip budget. Out of family the threshold still defers most items, but the flips it leaves are 0.0% where static deletion at 24 to 26 layers changes 1.5% to 3.8%. So the prediction holds once the gate is recalibrated, and the gain is small: about 7% of the layers of the static cut, before the cost of running the head at the exit, which we do not count. The raw top probability of a deleted-layer model is a poor gate because its scale has drifted (Section 11). The gate is still weak: the top probability at the 20-layer exit separates the items that later disagree with the full model from the rest with an AUROC of 0.64 in family and 0.78 on the public sets, against 0.90 and 0.81 for the full model's own margin, which is not available at the exit.

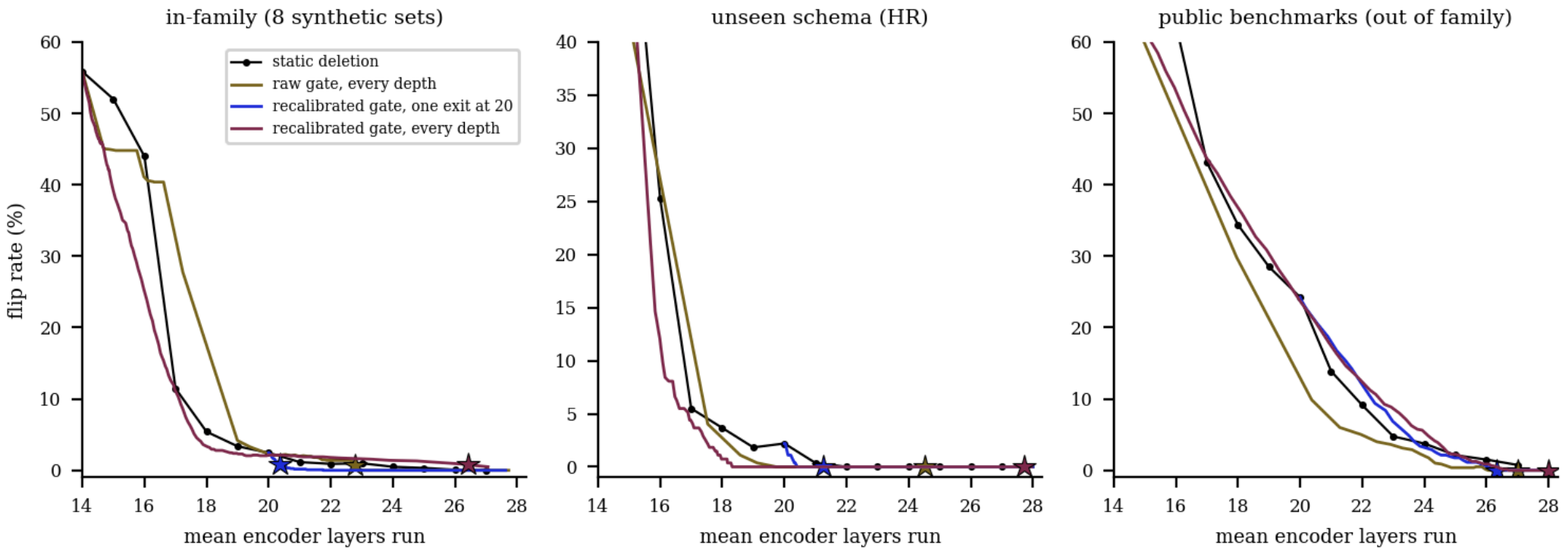

**Figure 17.** Mean encoder layers run against flip rate in the three item groups: static deletion of the top layers (black), and early-exit cascades over the fp32 head as the threshold varies (coloured curves; stars mark the threshold chosen on in-family items for at most 1% flips).

## 16 Repairing calibration with one temperature

Section 11 found that deleting layers leaves the ranking of items nearly intact while the reported probabilities move to a different scale, and left the cheapest repair untested. A deployment that thresholds on probability needs the numbers to be right, and a single temperature is the standard fix [14]. We test it for every layer-deleted model, with two ways to choose the temperature, on the nine synthetic sets by leaving one set out at a time. Fitting on labelled items is one. The other needs no labels: choose the temperature that makes the compressed model's probabilities closest, in KL divergence, to the fp32 model's on the same unlabelled items, which uses the model we are replacing as the teacher. We predicted that one scalar would bring the six-layer ECE under 15%.

**Table 19.** ECE (%) before and after one fitted temperature. 'Labels': a temperature fitted on the labelled items of the other synthetic sets, applied to the left-out set (public-benchmark sets use the temperature fitted on all nine synthetic sets). 'fp32': a temperature that makes the compressed probabilities match the fp32 model's on the other sets, with no labels at all. 'T' is the mean fitted temperature (labels). Decisions do not change.

| Recipe | T | ECE synthetic: raw | labels | fp32 | ECE public: raw | labels | fp32 |
|---|---|---|---|---|---|---|---|
| fp32 reference (no compression) | n/a | 5.9 | n/a | n/a | 20.0 | n/a | n/a |
| delete 2 layers | 1.21 | 4.8 | 3.7 | 4.9 | 14.5 | 10.8 | 14.7 |
| delete 4 | 0.83 | 6.2 | 5.6 | 4.3 | 3.1 | 4.1 | 8.2 |

| Recipe | T | ECE synthetic: raw | labels | fp32 | ECE public: raw | labels | fp32 |
|---|---|---|---|---|---|---|---|
| delete 6 | 0.24 | 39.2 | 2.8 | 3.6 | 7.5 | 9.0 | 11.3 |
| delete 8 | 0.09 | 55.5 | 2.5 | 3.6 | 10.7 | 7.3 | 9.7 |
| delete 10 | 0.03 | 59.6 | 4.9 | 4.1 | 9.0 | 9.8 | 11.0 |
| delete 12 | 0.02 | 34.8 | 28.4 | 28.4 | 4.8 | 5.8 | 5.8 |
| 4-bit groups | 1.19 | 5.0 | 2.8 | 4.2 | 17.6 | 14.2 | 16.6 |
| 3-bit groups | 1.14 | 4.3 | 3.2 | 3.3 | 19.3 | 16.5 | 15.2 |
| FFN 75% | 1.73 | 15.9 | 13.3 | 13.2 | 18.1 | 8.0 | 7.3 |
| 12 heads | 0.68 | 13.4 | 7.8 | 8.9 | 5.0 | 7.9 | 6.4 |
| 50% zeros | 0.70 | 9.8 | 4.8 | 4.9 | 6.6 | 10.2 | 10.2 |

It does, and by a wide margin. For the model with six layers deleted, ECE on the synthetic sets falls from 39.2% to 2.8% with labels and to 3.6% with the teacher-matched temperature, against 5.9% for fp32. At eight layers it falls from 55.5% to 2.5% and 3.6%. The fitted temperature is 0.24 at six layers and 0.09 at eight: the shortened model produces scores that are too flat, and the fix is to stretch them. The label-free choice is as good as the labelled one to within 1.3 points for the deleted-layer models up to ten layers, which means a team can repair the calibration of a layer-deleted model without collecting any labels. Three limits. On the unseen-schema set the repaired six-layer model still has ECE 14.8% (teacher-matched 7.1%) against 7.5% for fp32, so a temperature fitted on other schemas is not enough when the schema is new. On the public benchmarks the raw ECEs of the deleted-layer models were already lower than that of fp32 (7.5% against 20.0%) because fp32 is over-confident there (mean confidence 57%, accuracy 37%) and the flat scores offset that; stretching them back moves the ECE to 9.0%. And a model with twelve layers deleted is beyond a temperature (ECE 28.4% after fitting), because its decisions have already moved.

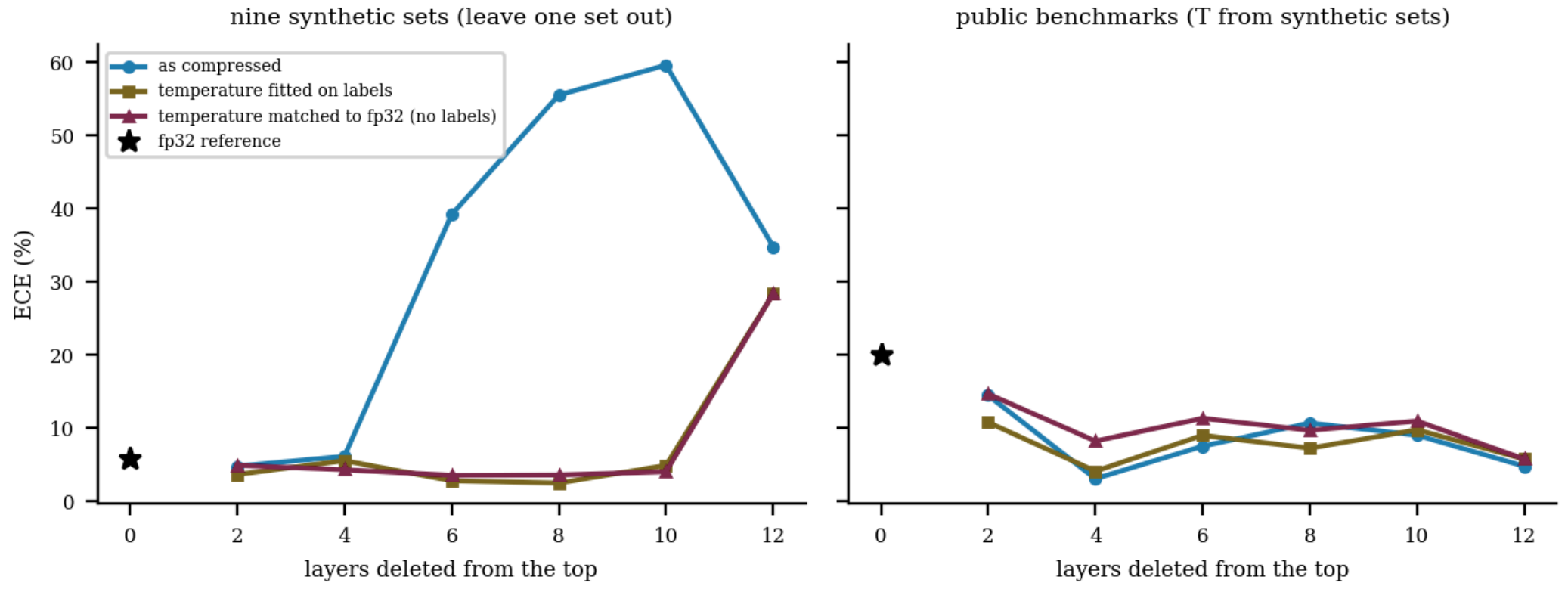

**Figure 18.** ECE of layer-deleted models before and after one fitted temperature, on the synthetic sets (left; leave-one-set-out) and on the public benchmarks (right).

## 17 Cost: FLOPs, bytes and CPU time

Cost has two kinds of measures. Parameters, stored bytes and FLOPs per item follow from the architecture and the question length, so they are the same on any machine, and they are the main cost axis of this study. Wall-clock time depends on the processor, the kernels and whatever else is running. The machine here was shared with other jobs and the GPU was busy for the whole period, so we took no GPU timings at all and treat the CPU timings in Table 20 as noisy context, not as results. FLOPs count 2 per weight per token for the encoder matrices of the layers that remain, plus the attention-mixing term, with the 128-token window on two of every three layers; at the mean test length of 109 tokens the window never binds, so local and global layers cost the same. Items were run one per question as the runtime does.

**Table 20.** Cost of the main configurations, fine-tuned checkpoint. Params, MiB and GFLOPs are machine independent: the stored model (fp16 for floating-point tensors; int8 plus fp16 scales for int8), and encoder FLOPs per item at the mean test length of 109 tokens. The CPU columns are noisy context, one run per configuration on a shared machine: median ms per item at batch 1 / 16 with 8 threads and the peak working set of the process in GiB (includes the Python runtime). The last column is the share of decisions changed against fp32 (n/a for combinations that were timed but not evaluated).

| Configuration | Params (M) | MiB | GFLOPs | FLOPs vs fp32 | CPU ms (b1 / b16) | CPU peak GiB | Changed % |
|---|---|---|---|---|---|---|---|
| fp32 | 421 | 804 | 76 | 1.00 | 132 / 114 | 2.9 | 0.0 |
| bf16 autocast | 421 | 804 | 76 | 1.00 | 816 / 782 | 3.5 | 0.2 |
| int8 dynamic, per-tensor | 421 | 452 | 76 | 1.00 | 80 / 57 | 6.0 | 30.4 |
| int8 dynamic, per-channel | 421 | 452 | 76 | 1.00 | 91 / 58 | 5.9 | 30.1 |
| drop top 4 layers | 372 | 710 | 65 | 0.86 | 110 / 97 | 2.9 | 0.4 |
| drop top 8 layers | 323 | 617 | 54 | 0.71 | 90 / 81 | 2.9 | 2.4 |
| drop top 14 layers | 250 | 476 | 38 | 0.50 | 75 / 63 | 2.9 | 56.1 |
| drop top 18 layers | 201 | 383 | 27 | 0.36 | 58 / 46 | 2.9 | n/a |
| int8 per-channel + drop 4 | 372 | 405 | 65 | 0.86 | 74 / 49 | 5.6 | n/a |
| int8 per-channel + drop 8 | 323 | 358 | 54 | 0.71 | 64 / 45 | 5.0 | n/a |
| int8 per-channel + drop 14 | 250 | 288 | 38 | 0.50 | 48 / 32 | 4.4 | n/a |
| FFN sliced to 50% | 308 | 588 | 51 | 0.68 | 100 / 84 | 2.9 | 60.6 |
| 8 of 16 heads | 363 | 692 | 63 | 0.82 | 106 / 93 | 2.9 | 56.4 |

### 17.1 Machine-independent cost

The full encoder needs 76 GFLOPs per item. Deleting layers scales this almost exactly with depth: eight layers removed leaves 71% of the FLOPs, fourteen leave 50%. Halving the FFN width saves 32% and using half the heads saves 18%. Quantisation and vocabulary trimming change bytes, not FLOPs. A recipe that stacks layer deletion with 6-bit weights and a trimmed vocabulary therefore moves two axes at once (Section 19). Table 20 lists compute and storage of the main configurations side by side.

### 17.2 CPU timings as noisy context

On the CPU, dynamic int8 (per-channel) took 2.0× less time than fp32 at batch 16 and 1.5× less at batch 1; deleting the top 8 layers (29% fewer FLOPs) gave 1.4×, the top 14 gave 1.8×, and int8 with 8 layers deleted gave 2.5×. Time fell to 0.86, 0.71, 0.56, 0.40 of the fp32 time with 4, 8, 14 and 18 layers deleted, against 0.86, 0.71, 0.50, 0.36 of the FLOPs, so on this CPU time follows FLOPs for layer deletion within the noise described below. bf16 autocast, the runtime default, was 6.9× *slower* than fp32 on this CPU, which has no native bf16 matrix path that this PyTorch build uses. These ratios come from one run each and the processor was 0-100% busy just before the runs (median 57%). Only ratios well above 1.3 should be read as real, and nothing below that.

### 17.3 How noisy are the timings?

The inter-quartile range inside a run is 4% to 17% of the median (median 7%), which understates the problem because it does not include other jobs starting and stopping between runs. The thread sweep shows the rest: for fp32 the batch-16 time in ms by thread count was 306 (1), 158 (2), 92 (4), 114 (8), 63 (16), 49 (24). The 8-thread run, the one used for every other configuration, took 114 ms against 92 ms with four threads, which more threads cannot cause: it reflects load from other jobs. The main-configuration timings carry the same kind of error. We did not repeat the runs while other work was active, because that would only measure the other work. A clean comparison needs an idle machine, several repeats per configuration and the spread over them; until then the FLOPs and bytes are the numbers to quote.

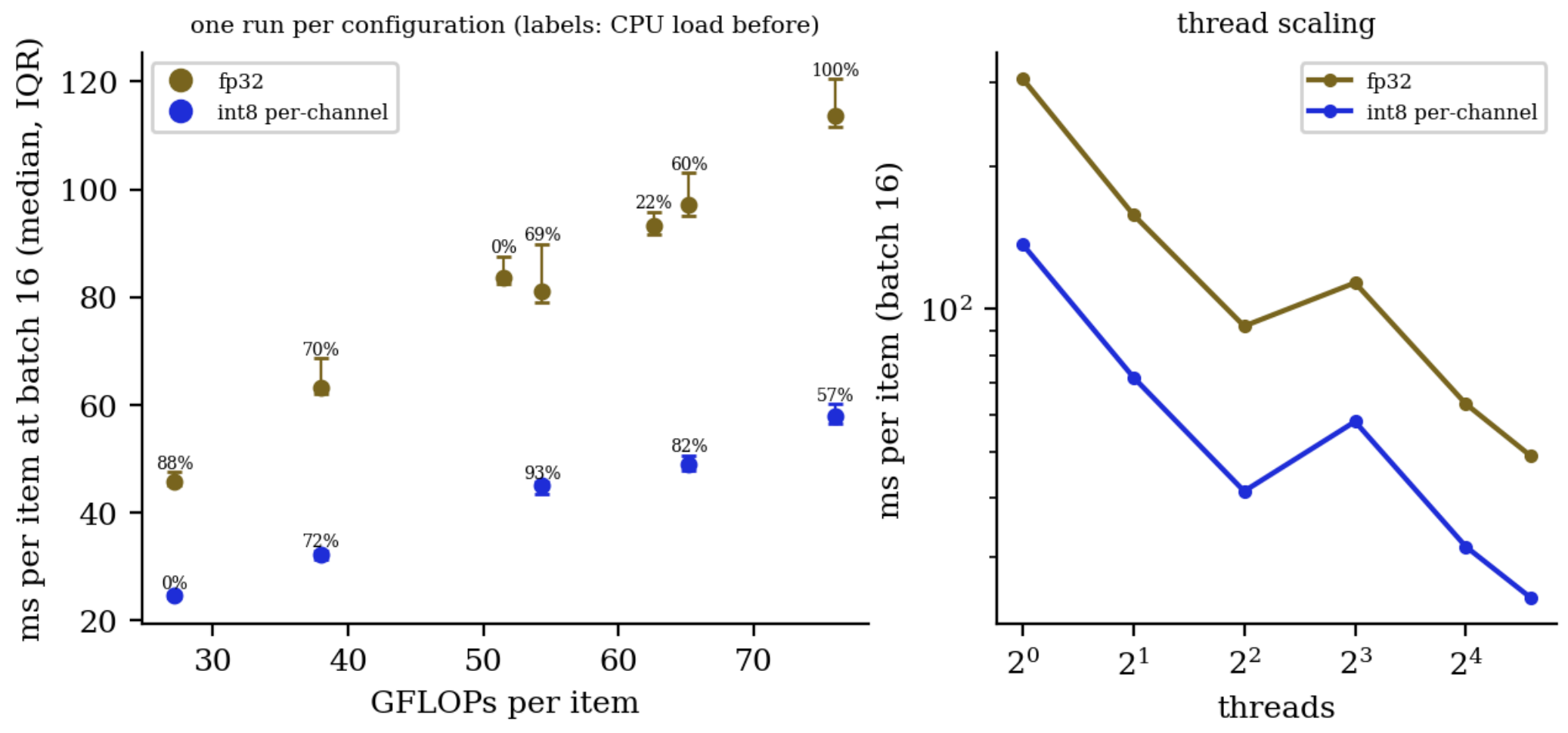

**Figure 19.** Left: measured batch-16 time against FLOPs (one run per configuration; the number above a point is the CPU load in percent just before the run, the bar is the inter-quartile range inside the run). Right: batch-16 time against the number of threads.

## 18 Retraining the shortened model

Layer deletion was the most promising training-free knob and it breaks down beyond about 9 layers. The standard way to buy back depth is to fine-tune the shortened model, as Sajjad et al. [3] and D K et al. [15] do, with patient distillation from the full model [26], or with a short healing run after pruning, as Gromov et al. [24] do for large language models. We ran it here for the top 8 and 12 layers, in two ways. In the first the student learns the stored labels. In the second it learns the fp32 model's own output distribution on the same 2,818 training rows, with no labels, so that it copies the model it replaces. Both use two epochs, learning rate 2e-05, an effective batch of 32 and frozen token embeddings; a run takes 2 to 10 minutes on a GPU shared with other jobs, with a peak of 5.3 GiB. Nothing from the test sets, the unseen schema or the public benchmarks is used. We expected distillation to bring in-family flips at 12 deleted layers below 10% and to beat training on labels. At 8 deleted layers (20 layers left) distillation raises in-family flips from 2.5% to 5.8% and training on labels to 7.0%, while in-family accuracy moves from 87.6% to 88.5% and 89.6%; public-benchmark flips go from 24.2% to 23.9%. A likely reading is that the untrained 20-layer model is already a close copy of the full one and two epochs on the training rows move weights that did not need moving, so retraining pays only for cuts deep enough to break the model; we did not test that reading.

**Table 21.** Retraining a shortened student. All students start from the fine-tuned checkpoint with the top layers removed and train for two epochs on its own training rows (in-family), with the embeddings frozen. 'Gold labels' trains on the stored labels; 'distilled' trains on the fp32 model's own output distribution and sees no labels. Accuracy is on the in-family items; flips are the % of decisions that differ from the 28-layer fp32 model; the last two columns give the ECE on the nine synthetic sets as the student stands and after one temperature matched to the fp32 model without labels (leave-one-set-out, as in the temperature section).

| Student | Layers | MiB | Acc. in-family % | Flips in-family % | Flips unseen schema % | Flips public % | ECE synthetic % | ECE after temperature % |
|---|---|---|---|---|---|---|---|---|
| fp32, 28 layers | 28 | 804 | 88.0% | 0.0 | 0.0 | 0.0 | 5.9 | 6.0 |
| delete 6, no retraining | 22 | 663 | 87.9% | 0.9 | 0.0 | 9.1 | 39.2 | 3.6 |
| delete 8, no retraining | 20 | 617 | 87.6% | 2.5 | 2.2 | 24.2 | 55.5 | 3.6 |
| delete 8, retrained on gold labels | 20 | 617 | 89.6% | 7.0 | 5.5 | 24.8 | 27.3 | 1.8 |
| delete 8, retrained on distilled from fp32 | 20 | 617 | 88.5% | 5.8 | 5.5 | 23.9 | 27.8 | 2.4 |
| delete 12, no retraining | 16 | 523 | 54.6% | 44.0 | 25.3 | 62.5 | 34.8 | 28.4 |

| Student | Layers | MiB | Acc. in-family % | Flips in-family % | Flips unseen schema % | Flips public % | ECE synthetic % | ECE after temperature % |
|---|---|---|---|---|---|---|---|---|
| delete 12, retrained on gold labels | 16 | 523 | 88.1% | 9.7 | 15.4 | 44.4 | 27.4 | 2.8 |
| delete 12, retrained on distilled from fp32 | 16 | 523 | 87.4% | 8.6 | 11.4 | 45.0 | 31.9 | 1.9 |

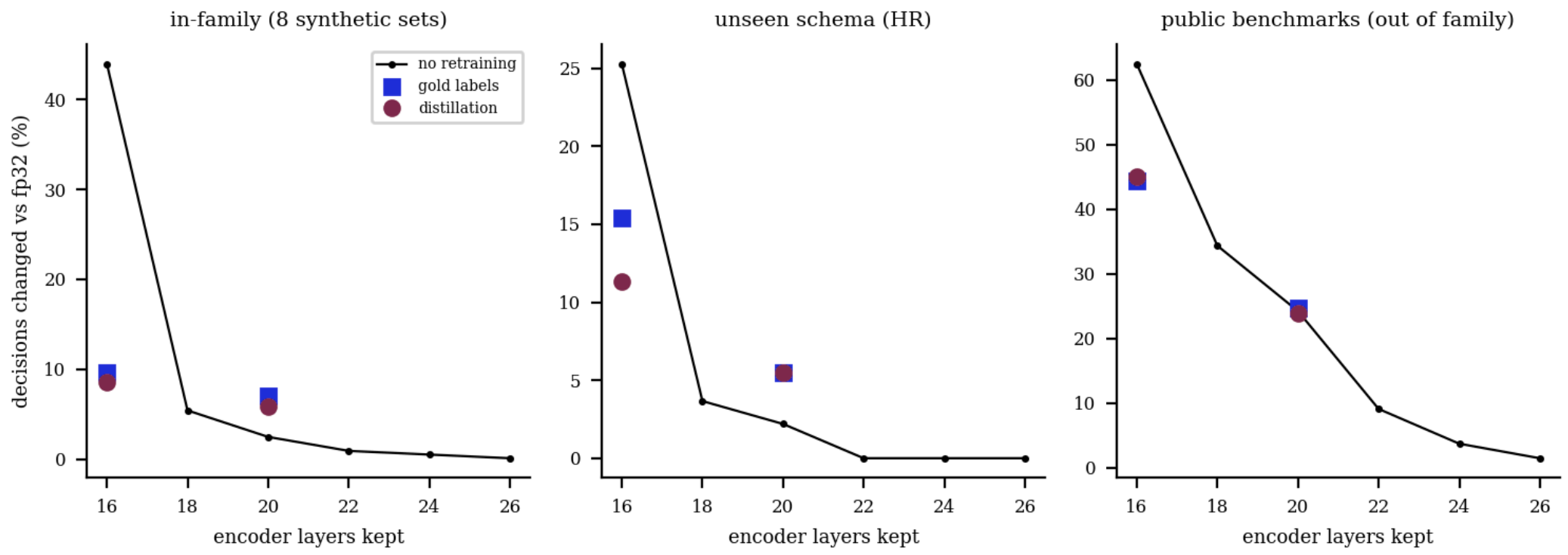

**Figure 20.** Decisions changed against the 28-layer fp32 model for layer-deleted students: no retraining (line) and two epochs of retraining on gold labels (squares) or on the fp32 model's output distribution (circles); one panel per item group.

Both ways recover the accuracy of the full model in family (87.4% distilled, 88.1% on labels, against 88.0% for fp32 and 54.6% for the untrained cut) and cut the in-family flips at 12 deleted layers from 44.0% to 8.6% and 9.7%. Distillation is ahead by 1.1 points in family (paired McNemar p = 0.17), and further ahead on the unseen schema: 11.4% against 15.4% flips and 81.3% against 75.8% accuracy (fp32: 84.2%; untrained cut: 73.6%; p = 0.043 for the flips). A student that copies the original's output distribution keeps more of its decisions than one that learns the labels, which is the difference between a smaller model that is as good and a smaller model that is the same. The prediction that distillation brings the in-family flips under 10% and beats labels holds, although the margin over labels is not significant in family.

Out of family retraining does little. On the public benchmarks the 12-layer students change 45.0% (distilled) and 44.4% (labels) of decisions against 62.5% for the untrained cut, so most of what retraining buys is confined to the family it was trained on. We trained on in-family rows only; whether training on other kinds of text would move the out-of-family number is not tested. Calibration is not repaired either: the ECE of the students on the synthetic sets is 31.9% and 27.4%, because neither loss touches the temperature, but the one-temperature repair of Section 16 brings the distilled student to 1.9% (Table 21).

A healed student can also serve as the first stage of a cascade whose second stage is the full model. Because the two models share no computation, a deferred item costs the 16 layers of the student plus all 28 of the full model. With a recalibrated gate the cheapest cascade at 1% flips in family runs 25.7 layers on average against 22 for static deletion of the top layers, so it saves nothing: retraining gives a smaller standalone model, not a cheaper gate.

What was checked earlier remains: the in-memory deletion stands for a shipped file. The standalone 16-layer checkpoint, written with its tensors renumbered and loaded by the unmodified runtime, differs from the in-memory deletion of the same layers on 0 of 1,494 items, so the in-memory sweeps in this report stand for what a shipped, shorter file would do. The 16-layer file is 523 MiB for the untrained cut and the retrained students alike.

**Table 22.** Layer deletion, in memory and as a standalone checkpoint (fine-tuned checkpoint, no retraining). 'Decisions changed' is against the 28-layer fp32 model.

| Checkpoint | Layers | Params (M) | MiB | Accuracy [95% CI] | Decisions changed vs 28-layer fp32 | ΔECE (points) | ECE % |
|---|---|---|---|---|---|---|---|

| Checkpoint | Layers | Params (M) | MiB | Accuracy [95% CI] | Decisions changed vs 28-layer fp32 | ΔECE (points) | ECE % |
|---|---|---|---|---|---|---|---|
| 28 layers, fp32 (reference) | 28 | 421 | 804 | 87.3% [85.5-88.9] | n/a | n/a | 5.80 |
| 20 layers, deleted in memory | 20 | 323 | 617 | 86.9 [85.1-88.5] | 36 (2.41%) [1.75-3.32] | +49.71 [+46.24, +52.32] | 55.51 |
| 16 layers, deleted in memory | 16 | 274 | 523 | 58.1 [55.6-60.6] | 606 (40.56%) [38.10-43.07] | +28.99 [+25.17, +31.73] | 34.79 |
| 16 layers, physically removed checkpoint | 16 | 274 | 523 | 58.1 [55.6-60.6] | 606 (40.56%) [38.10-43.07] | +28.99 [+25.17, +31.73] | 34.79 |

## 19 Pareto frontiers, matched sizes and a recipe card

Figure 21 plots every configuration we evaluated on the fine-tuned checkpoint (110 of them, excluding the single-tensor and single-layer probes) by stored size and by the share of decisions that agree with fp32. The frontier (configurations that no other configuration beats on both axes) is made of combo, drop, vocab, wq configurations. The cheapest frontier point with at least 99% agreement is 'combo_d6_q8' at 383 MiB, against 804 MiB for the unmodified fp16 file. That point deletes layers, so it carries the large calibration drift described in Section 11; the frontier here is on decisions only.

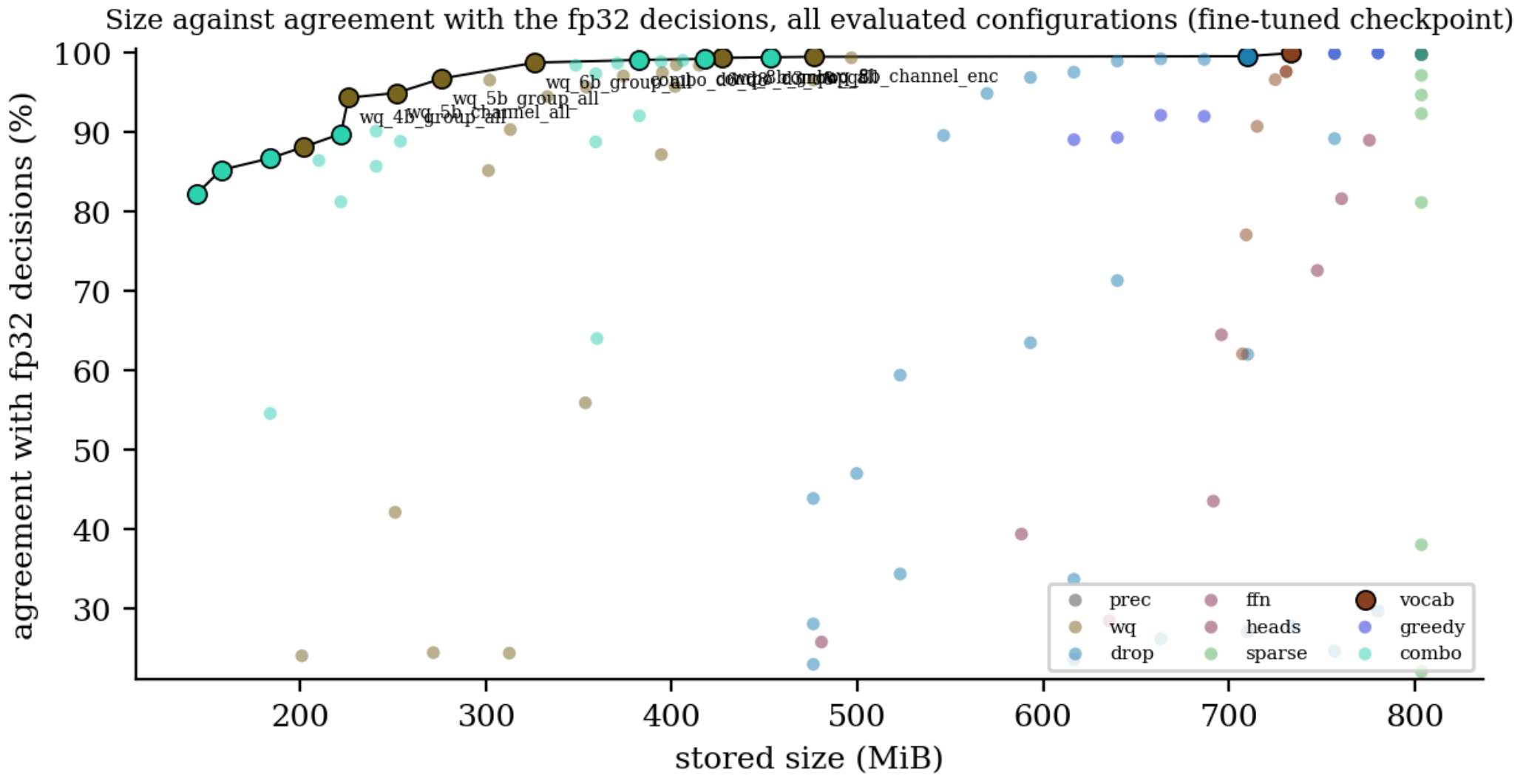

**Figure 21.** Every evaluated configuration (fine-tuned checkpoint): stored size against the share of decisions that agree with fp32. Black outline and line: size-agreement frontier. Colours are knob families.

The same plot with encoder FLOPs per item in place of size gives a frontier made of drop configurations. Size and FLOPs do not coincide: rounding weights and trimming the vocabulary shrink the file without changing the FLOPs, while deleting layers or slicing the FFN shrink both, so FLOPs are the right axis when the cost that matters is compute, not storage. The recipe table below lists both.

### 19.1 Recipe card

**Table 23.** Recipe card for the fine-tuned checkpoint (1,494 items). Tier A: upper 95% bound of the changed-decision rate ≤ 1.5% and the ΔECE interval inside ±1 point; tier B: point estimate ≤ 5%; tier C: above. '*': on the size-agreement frontier of all 110 evaluated configurations. 'GFLOPs': encoder FLOPs per item (Section 17).

| Recipe | Params (M) | MiB | GFLOPs | Accuracy [95% CI] | Decisions changed | ΔECE (points) | Tier |
|---|---|---|---|---|---|---|---|
| fp16 weights | 421 | 804 | 76 | 87.3 [85.6-88.9] | 2 (0.13%) [0.04-0.49] | -0.12 [-0.50, +0.19] | **A** |
| 8-bit, per-channel | 421 | 477 | 76 | 87.5 [85.7-89.1] | 7 (0.47%) [0.23-0.96] | +0.13 [-0.67, +0.71] | **A** * |

| Recipe | Params (M) | MiB | GFLOPs | Accuracy [95% CI] | Decisions changed | ΔECE (points) | Tier |
|---|---|---|---|---|---|---|---|
| 6-bit, groups of 32 | 421 | 415 | 76 | 87.6 [85.8-89.1] | 23 (1.54%) [1.03-2.30] | +0.13 [-0.96, +0.89] | **B** |
| 5-bit, groups of 32 | 421 | 374 | 76 | 86.8 [85.0-88.4] | 43 (2.88%) [2.14-3.85] | +0.43 [-0.69, +1.26] | **B** |
| 4-bit, groups of 32 | 421 | 333 | 76 | 85.8 [83.9-87.5] | 82 (5.49%) [4.44-6.76] | -0.43 [-1.47, +0.65] | **C** |
| drop top 4 layers | 372 | 710 | 65 | 87.5 [85.7-89.1] | 6 (0.40%) [0.18-0.87] | +0.30 [-2.13, +2.36] | **B** * |
| drop top 6 layers | 348 | 663 | 60 | 87.2 [85.4-88.8] | 11 (0.74%) [0.41-1.31] | +33.33 [+30.18, +35.91] | **B** |
| drop top 7 layers | 335 | 640 | 57 | 87.3 [85.5-88.9] | 15 (1.00%) [0.61-1.65] | +46.19 [+42.76, +48.92] | **B** |
| greedy: delete 6 layers | 348 | 663 | 60 | 85.4 [83.5-87.1] | 117 (7.83%) [6.57-9.30] | -1.60 [-3.49, +0.49] | **C** |
| greedy: delete 8 layers | 323 | 617 | 54 | 83.6 [81.6-85.4] | 163 (10.91%) [9.43-12.59] | +3.13 [+0.72, +5.43] | **C** |
| vocabulary 5,000 ids | 375 | 715 | 76 | 80.3 [78.2-82.3] | 138 (9.24%) [7.87-10.81] | +3.60 [+2.00, +4.80] | **C** |
| 8-bit channel (bf16 weights) | 421 | 453 | 76 | 87.3 [85.5-88.9] | 8 (0.54%) [0.27-1.05] | -0.02 [-0.79, +0.67] | **A** * |
| drop 4 + 8-bit | 372 | 406 | 65 | 87.0 [85.2-88.6] | 14 (0.94%) [0.56-1.57] | -0.14 [-2.54, +2.06] | **B** |
| drop 6 + 8-bit | 348 | 383 | 60 | 87.0 [85.2-88.6] | 13 (0.87%) [0.51-1.48] | +33.22 [+30.01, +35.84] | **B** * |
| drop 6 + 6-bit + vocab 5k | 301 | 241 | 60 | 80.9 [78.8-82.8] | 147 (9.84%) [8.43-11.45] | +30.24 [+26.67, +33.06] | **C** |
| drop 8 + 6-bit + vocab 5k | 277 | 222 | 54 | 80.7 [78.6-82.6] | 152 (10.17%) [8.74-11.81] | +44.51 [+41.11, +47.35] | **C** * |
| 4-bit + vocab 5k | 375 | 210 | 76 | 78.7 [76.6-80.7] | 202 (13.52%) [11.88-15.35] | +2.54 [+0.92, +4.07] | **C** |
| drop 8 + 4-bit + vocab 5k | 277 | 158 | 54 | 77.8 [75.7-79.9] | 220 (14.73%) [13.02-16.61] | +42.35 [+39.08, +45.10] | **C** * |

**How to read it.** The smallest tier-A recipe is '8-bit channel (bf16 weights)': 453 MiB, a 44% reduction from the 804 MiB fp16 file, with 0.54% [0.27-1.05] of decisions changed. The smallest tier-B recipe is '5-bit, groups of 32' at 374 MiB (2.9% [2.1-3.9] changed). Tier A means we could not distinguish the compressed model from the original on 1,494 items; it does not mean identical behaviour on every input, and it was selected after looking at these items, so the same recipe should be confirmed on a fresh sample before it is trusted (300 fresh items with no flips would bound the flip rate below 1% at roughly 95% confidence). Recipes that delete layers can show large ΔECE in the table even when few decisions change; their tier B rests on decisions alone and they would need recalibration (Section 11).

**Quantization or pruning at the same size?** Kuzmin et al. [23] compare the two on accuracy for eight large models and find that quantization is better in most cases, with pruning ahead only at very high compression. We ask the same question with decision flips as the outcome and stored size as the budget (Table 24, Figure 22). Weight rounding gives the lowest flip rate in 3 of the 5 size classes, and the combination of four deleted layers with 8-bit rounding in 2. Layer deletion is competitive only close to the full size: at 600 MiB its best configuration changes 11.9% of decisions against 1.5% for rounding, and at 500 MiB 58.3%. FFN and head slicing are far behind at every size. This agrees with Kuzmin et al. for this encoder, with the caveats that our rounding is fake quantisation (the file size is counted, no integer kernel runs), that deletion and slicing receive no retraining here, and that the configurations are the ones we chose to evaluate and not a search.

**Table 24.** The lowest flip rate (all 2,294 items; configuration in brackets) among the evaluated configurations of each knob family whose stored size is at most the limit (MiB). Sparsity is left out because it does not shrink the stored size in this accounting; precision alone stays at 804 MiB.

| Size limit (MiB) | weight rounding | layer deletion | FFN slicing | head slicing | combination |
|---|---|---|---|---|---|
| 700 | 1.5% (wq_8b_channel_enc) | 2.2% (drop_top5) | 45.3% (ffn_0.75) | 59.1% (heads_8) | 2.5% (combo_d4_q8) |
| 600 | 1.5% (wq_8b_channel_enc) | 11.9% (drop_top9) | 65.2% (ffn_0.5) | none fits | 2.5% (combo_d4_q8) |
| 500 | 1.5% (wq_8b_channel_enc) | 58.3% (drop_top13) | none fits | none fits | 2.5% (combo_d4_q8) |
| 450 | 3.1% (wq_6b_group_enc) | none fits | none fits | none fits | 2.5% (combo_d4_q8) |
| 400 | 5.1% (wq_6b_channel_enc) | none fits | none fits | none fits | 4.3% (combo_d6_q8) |

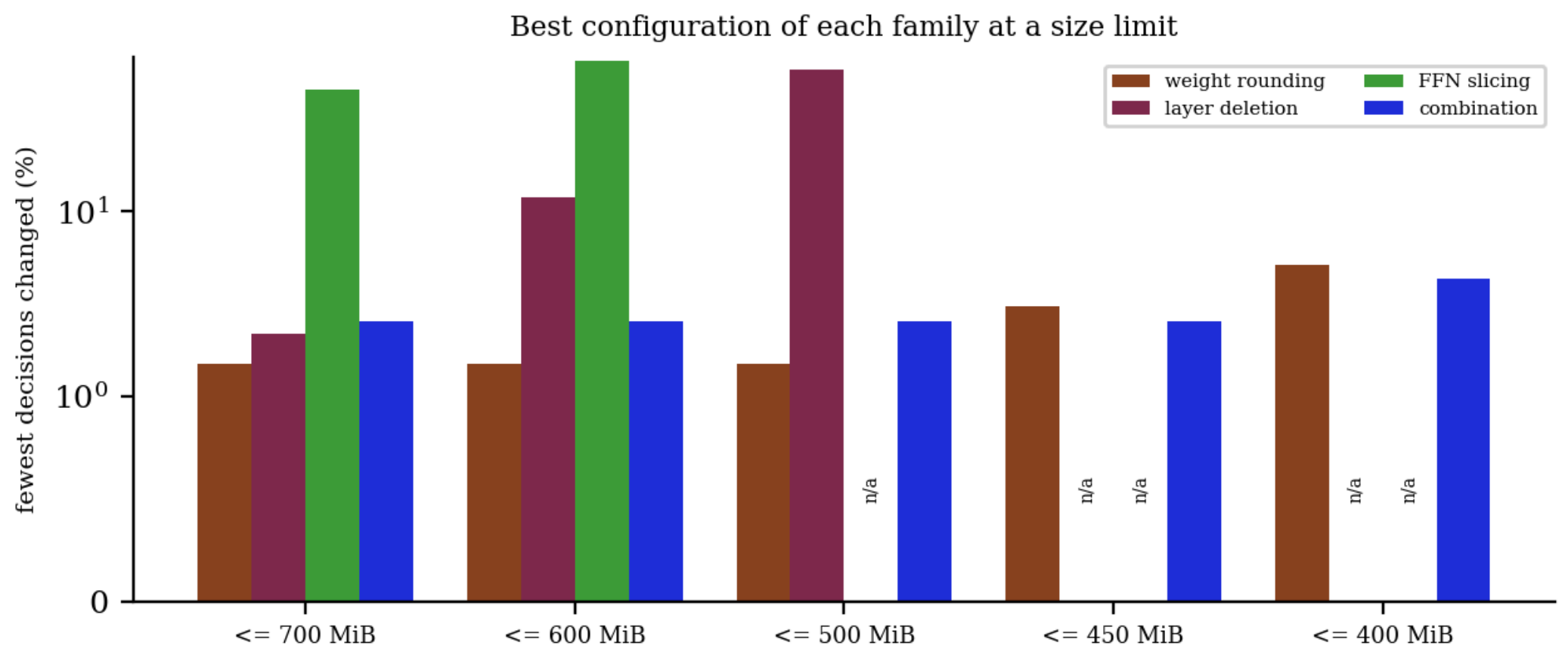

**Figure 22.** Lowest flip rate among the configurations of each knob family within a size limit (symmetric-log axis; fine-tuned checkpoint, all items).

## 20 Against a TF-IDF classifier

The strongest argument against shrinking a 421-million-parameter encoder is that a much smaller model may already do the job. A bag-of-n-grams classifier is the standard test. Table 25 and Figure 23 compare TF-IDF features (word 1-2-grams plus character 2-5-grams) with logistic regression, trained per task on exactly the rows used for that task, with the fine-tuned and base Laya checkpoints on the 1,008 items of the six tasks for which a per-task classifier exists.

**Table 25.** TF-IDF + logistic regression (one classifier per task, trained on that task's training rows only) against the fine-tuned and base Laya checkpoints on identical items. 'Diff' is fine-tuned minus TF-IDF in accuracy points with an exact McNemar p-value. Last two columns: serialised classifier size in MiB and its median CPU time per item (one thread).

| Task | Items | Train rows | TF-IDF % | Fine-tuned % | Diff (p) | Base % | Compressed % | TF-IDF MiB | TF-IDF ms |
|---|---|---|---|---|---|---|---|---|---|
| JSON tickets | 58 | 233 | 84.5 | 89.7 | +5.2 (p=0.508) | 74.1 | 87.9 | 1.3 | 1.26 |
| hardware failure | 82 | 544 | 87.8 | 96.3 | +8.5 (p=0.065) | 62.2 | 89.0 | 1.2 | 0.92 |
| 12-way topic | 195 | 1145 | 88.2 | 92.8 | +4.6 (p=0.078) | 73.8 | 89.2 | 4.4 | 1.56 |
| multilingual | 322 | 1707 | 91.9 | 73.9 | -18.0 (p<0.001) | 61.5 | 71.7 | 4.2 | 1.40 |
| urgency (ordinal) | 78 | 148 | 91.0 | 88.5 | -2.6 (p=0.774) | 57.7 | 85.9 | 0.5 | 0.88 |
| HR (unseen schema) | 273 | 443 | 90.1 | 84.2 | -5.9 (p=0.033) | 82.8 | 67.8 | 2.1 | 1.21 |

Pooled over these tasks the TF-IDF classifiers score 89.9% [87.9-91.6] and fine-tuned Laya 84.2% [81.8-86.3] (77.5% for the compressed recipe 'drop 6 layers + 6-bit + vocabulary'). Fine-tuned Laya is significantly better (McNemar $p<0.05$) on: none; it is lower on: multilingual, urgency (ordinal), HR (unseen schema); and the rest are statistical ties at these sample sizes (JSON tickets, hardware failure, 12-way topic). The base checkpoint is below TF-IDF on most tasks, which is the reason to fine-tune at all. The comparison is not level: the fine-tuned checkpoint was trained on the 2,818-row general set, which has 399 multilingual and 148 urgency rows and no HR rows, whereas each classifier was trained on its own task's rows (1,707 multilingual, 443 HR). That is why Laya trails most on multilingual and HR, and it means this table says little about a Laya fine-tuned on those tasks.

**On cost the classifier wins by orders of magnitude.** A per-task TF-IDF model is 2.3 MiB on average and takes 1.20 ms per item on one CPU thread, against 804 MiB and 132 ms for the fp32 Laya (a 110× gap in time, measured with eight threads for Laya), and 358 MiB and 64 ms for the most compressed CPU configuration we timed (int8 with eight layers deleted), still 54× slower. Two things are not captured by the table. First, the classifier needs training rows for every new question; Laya was fine-tuned once and one model answers all of these tasks, taking the question and options as input. Second, we did not measure the classifier's calibration or its sensitivity to wording, so the table compares accuracy and cost only.

**What this means for compression.** Compression narrows the cost gap only modestly: the cheapest Laya configuration we timed on the CPU is still 54 times slower and 279 times larger than a classifier. The case for the encoder is generality across questions, schemas and languages with one artefact, and for that case compression is worthwhile only if the decisions survive, which is what the earlier sections measure. Where one narrow, stable question is asked at volume and a few points of accuracy are acceptable, the classifier needs no compression study at all.

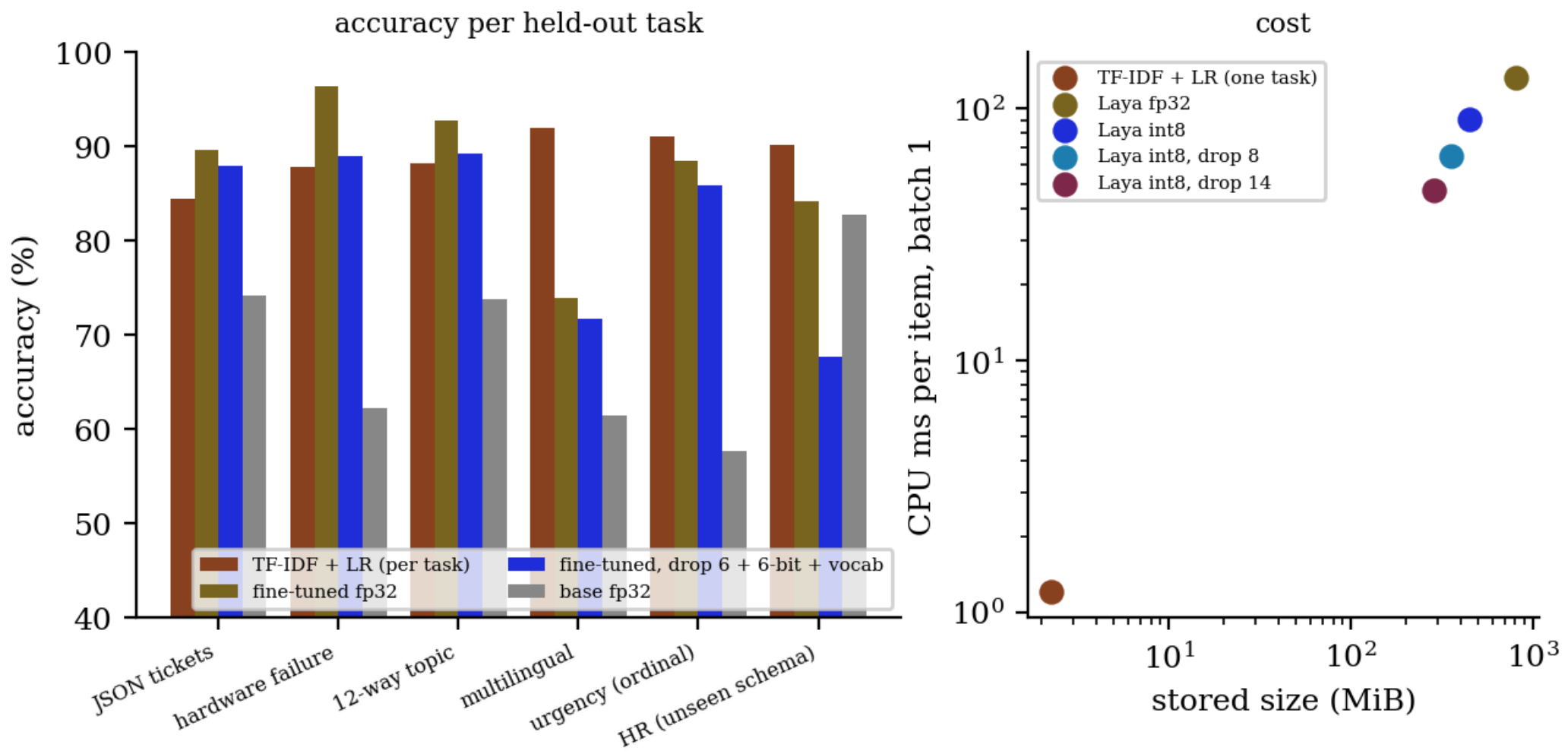

**Figure 23.** Left: accuracy of TF-IDF + logistic regression, fine-tuned Laya (fp32 and one compressed recipe) and base Laya on six tasks. Right: stored size and single-item CPU time of the classifier and of Laya variants (log axes).

## 21 Limitations and threats to validity

- **Synthetic, LLM-labelled data.** Labels were produced by one generator model and filtered by agreement with a second pass of the same model; an independent labeller agrees on most but not all test items. Accuracy numbers are consistency with that labelling, not with human judgement. Decision flips against fp32 do not depend on labels, which is why they carry the argument, but the *items* are synthetic: real text may be longer, noisier and less templated, and flips may concentrate differently.
- **One backbone, a few checkpoints.** Every checkpoint shares a 28-layer, 1,024-wide ModernBERT-large encoder. Nothing here tells us how a smaller or differently built Laya behaves. The seed and no-shuffle replications use the same training rows as the main checkpoint, so they measure the spread over retraining and not over data (Section 13).

- **Fake quantisation is not int8 inference.** The rounding sweeps measure the effect on decisions exactly but not the speed. Dynamic int8 was run through PyTorch only, not through the ONNX, OpenVINO or MNN exports that community builds use, so we make no statement about those builds.
- **Timing noise, and no GPU timings.** Other jobs shared the CPU and the GPU for the whole study (Section 17). CPU timings are single runs, absolute milliseconds are specific to this machine and moment, and no GPU latency was measured because the GPU was occupied. Only FLOPs, bytes and parameters are machine independent. FLOPs omit the option head, scorer and act head and ignore memory traffic.
- **Selection on the test items.** The recipe card, tier boundaries and the choice of representative recipes were made after seeing results on the 1,494 synthetic items; the greedy layer search is the only selection done on training rows. The early-exit threshold is chosen on in-family items and the temperature is fitted leaving one set out, so those two do not use the items they are scored on, but the list of configurations we evaluated was chosen by us. Intervals describe sampling error on these items and do not include the researcher's degrees of freedom. The recipes should be re-checked on fresh items.
- **The out-of-family rows are a stress test.** The 800 public-benchmark questions are recast as one-question rows with letter option names, a format the encoder never saw, and the model is near chance on them. They show that decisions are unstable where the model is unsure. They do not show how a compressed model behaves on a deployment's own out-of-family traffic, and flips on them are not losses in accuracy.
- **Stored probabilities.** The probabilities in the extended sweep are stored with four decimals. Margins use them with a floor of $10^{-4}$ (so a margin is at most about 9.2) and the TV values are computed from unrounded numbers before storage. The noise-floor runs use one GPU, one library version and one engine.
- **Small sets.** The ordinal set has 78 items and the JSON set 58; per-set flip rates for them have wide intervals, and statements about 'ordinal' and 'many-option' fragility are bounded by that.
- **Limited retraining.** Healing was two epochs on the in-family training rows with one learning rate, one seed per student and frozen token embeddings, at the depths in Table 21 (8 and 12 deleted layers). Quantisation-aware training, layer dropout, layer-wise distillation and healing on out-of-family data were not tried, so the healed numbers are a floor on what training can recover.
- **Cost accounting.** Early-exit cost counts encoder layers run and ignores the option head at each exit; the FLOP counts in Section 17 omit the head too.
- **Single-layer deletion is partial.** Only layers 0 to 3 were deleted one at a time; the per-layer picture comes from 3-bit rounding of one layer.
- **Not evaluated:** long inputs, adversarial inputs, throughput under concurrent load, energy, other devices or runtimes.

## 22 Conclusion

For a one-pass decision encoder of this size, the safe moves are half precision, per-channel or grouped 8-bit weights and a handful of top layers, and they stop well before the cost of the model approaches that of a bag-of-words classifier. The sharp edges are specific: single-scale rounding below six bits, activation quantisation of the feed-forward output projection, slicing feed-forward units or heads, deleting low layers, and anything judged only by accuracy or only in family. The probabilities drift before the decisions do, and when the layers that were deleted are the top ones, one fitted temperature repairs the probabilities with no labels. On items from outside the training family the same recipes change several times more decisions, because the model is unsure there and a small shift reorders its top two options. Total variation tracks flips for weight rounding and not for layer deletion, margins shift by a constant instead of shrinking, and an early exit beats a static cut only when its confidence gate is recalibrated first. A short distillation run brings the model with 12 layers deleted from 44% to 9% changed in-family decisions, with little effect out of family (45% changed there). At 8 deleted layers the same run raises in-family flips from 2.5% to 5.8%, so retraining pays only for cuts that break the model. What remains open is quantisation-aware training and layer dropout, how the community int8 exports behave on the same items, and whether the picture holds on real text, on a deployment's own traffic, and on idle-machine timings.

## 23 Cited works

## Appendix A  Per-set results for the recipes

Accuracy / decisions-changed (both in %) per test set for eight recipes; the third column is fp32 accuracy.

**Table 26.** Per-set accuracy and flips for representative recipes (fine-tuned checkpoint).

| Set | n | fp32 acc. | bf16 autocast | 8-bit channel | 6-bit group | 4-bit group | drop top 4 | drop top 7 | drop 4 + 8-bit | drop 6 + 6-bit + vocab |
|---|---|---|---|---|---|---|---|---|---|---|
| mixed probe | 224 | 91.1 | 91.1 / 0.0 | 91.1 / 0.4 | 91.1 / 2.2 | 90.6 / 2.7 | 91.5 / 0.9 | 91.1 / 1.8 | 90.6 / 1.8 | 87.1 / 5.8 |
| sentiment | 106 | 95.3 | 95.3 / 0.0 | 95.3 / 0.0 | 95.3 / 0.0 | 95.3 / 1.9 | 95.3 / 0.0 | 96.2 / 0.9 | 94.3 / 0.9 | 86.8 / 8.5 |
| JSON tickets | 58 | 89.7 | 89.7 / 0.0 | 89.7 / 0.0 | 89.7 / 0.0 | 89.7 / 3.4 | 89.7 / 0.0 | 89.7 / 0.0 | 87.9 / 1.7 | 87.9 / 5.2 |
| hardware failure | 82 | 96.3 | 96.3 / 0.0 | 96.3 / 0.0 | 96.3 / 0.0 | 96.3 / 1.2 | 96.3 / 0.0 | 96.3 / 0.0 | 96.3 / 0.0 | 89.0 / 7.3 |
| 12-way topic | 195 | 92.8 | 92.8 / 0.0 | 92.8 / 0.0 | 93.3 / 0.5 | 89.2 / 3.6 | 92.8 / 0.0 | 92.8 / 0.0 | 92.8 / 0.0 | 89.2 / 5.6 |
| multilingual | 322 | 73.9 | 73.6 / 0.3 | 74.2 / 0.6 | 75.2 / 2.5 | 71.4 / 8.1 | 74.5 / 0.6 | 73.6 / 1.6 | 73.6 / 1.2 | 71.7 / 7.8 |
| urgency (ordinal) | 78 | 88.5 | 91.0 / 2.6 | 91.0 / 2.6 | 88.5 / 2.6 | 91.0 / 5.1 | 88.5 / 2.6 | 87.2 / 3.8 | 88.5 / 2.6 | 85.9 / 10.3 |
| code-mixed | 156 | 96.2 | 96.2 / 0.0 | 96.8 / 0.6 | 96.2 / 0.0 | 96.8 / 3.2 | 96.2 / 0.0 | 96.8 / 0.6 | 96.8 / 0.6 | 89.7 / 7.7 |
| HR (unseen schema) | 273 | 84.2 | 84.2 / 0.0 | 83.9 / 0.4 | 83.9 / 2.6 | 81.0 / 10.6 | 84.2 / 0.0 | 84.2 / 0.4 | 83.9 / 0.4 | 67.8 / 22.0 |

## Appendix B  Frozen inputs and code

Every number is recomputed at build time from these frozen inputs (SHA-256 prefix shown). Commands, from the project directory: `python a9_run.py v1 prec,wq,comp,layerq,drop,layer1,struct,vocab,greedy,combo` and the same for `base`; `python a9_cost.py drive cpu`; `python a9_refit.py make …` and `python a9_refit_eval.py …`; `python a9_tfidf_cost.py` (system Python with scikit-learn); the second round: `python a9b_data.py`, `python a9_ext.py v1 full`, `python a9_ext.py base core` and `python a9_ext.py <checkpoint> seed`, `python a9_heal.py kd_d12` (also hard_d12, kd_d8, hard_d8, kd_d6), `python a9_noise.py`. The analyses are in `a9x_lib.py` and `a9x_figs.py`; the report is built by `lp_u9.py`. The text was drafted and edited with LLM assistance; all numbers are recomputed from the stored results by the build scripts and were checked by the authors.

**Table 27.** Frozen inputs.

| File | Content | SHA-256 (first 16 hex) |
|---|---|---|
| `results/a9_grid_v1.json` | per-item answers of every GPU configuration, fine-tuned checkpoint | `c1cbd0c949f93bc3` |
| `results/a9_grid_base.json` | the same for the base checkpoint | `e8ead7fd3e96c9af` |
| `results/a9_ext_v1.json` | second-round sweep with full probabilities, 13 item sets (9 synthetic, 4 public benchmarks), fine-tuned checkpoint | `39dbeb7be40f5255` |
| `results/a9_ext_base.json` | second-round sweep, base checkpoint | `a087a2a3af13b424` |
| `results/a9_ext_v1_s1.json` | second-round sweep, other checkpoints (seeds, no shuffling, multilingual; files a9_ext_*.json) | `914902fe32c1bb2b` |
| `results/a9_heal_kd_d12.json` | healed students (files a9_heal_*.json) | `fe23b022f28a7586` |
| `results/a9_noise.json` | noise floor runs | `ca39ea849943ebc4` |

| File | Content | SHA-256 (first 16 hex) |
|---|---|---|
| `results/a9_cost_cpu.json` | CPU timings, memory, int8 evaluations | `0ae0e26b3a93b994` |
| `results/a9_tfidf.json` | TF-IDF + logistic regression: accuracy, size, latency | `ebdaacec63bce742` |
| `results/a9_refit_eval.json` | the standalone 16-layer checkpoint evaluated on the same items | `7b9f0f2c1ebc0494` |
| `results/xlabel_gemma4-31b.json` | independent labeller | `28f1fa32695a39f6` |
| `data/train.jsonl` | training rows | `18a3c345b94ac8e3` |
| `data/a9b_arc_e.jsonl` | public-benchmark rows (also a9b_arc_c, a9b_mmlu, a9b_hella), built by a9b_data.py with seed 9 | `f652b165ae6eef8e` |